\documentclass[sigconf]{acmart}

\copyrightyear{2022}
\acmYear{2022}
\setcopyright{acmlicensed}\acmConference[KDD '22]{Proceedings of the 28th ACM SIGKDD Conference on Knowledge Discovery and Data Mining}{August 14--18, 2022}{Washington, DC, USA}
\acmBooktitle{Proceedings of the 28th ACM SIGKDD Conference on Knowledge Discovery and Data Mining (KDD '22), August 14--18, 2022, Washington, DC, USA}
\acmPrice{15.00}
\acmDOI{10.1145/3534678.3539477}
\acmISBN{978-1-4503-9385-0/22/08}
\usepackage{amsmath,amsfonts}
\usepackage{euler}
\usepackage{algorithmic}
\usepackage{graphicx}
\usepackage{textcomp}
\usepackage{xcolor}
\usepackage{xspace}
\usepackage[ruled, vlined, linesnumbered]{algorithm2e}
\usepackage{algorithmic}
\usepackage{graphicx}
\usepackage{subcaption}
\usepackage{amsthm}
\usepackage{mathrsfs}
\usepackage{booktabs}
\usepackage{color}
\newcommand{\ourm}{\textsc{ProActive}\xspace}
\newcommand{\ours}{\textsc{ProAct}\xspace}

\newcommand{\cm}[1]{\mathcal{#1}}

\newcommand{\bs}[1]{\boldsymbol{#1}}
\newcommand{\eg}{\emph{e.g.}}
\newcommand{\ie}{\emph{i.e.}}
\newcommand{\etc}{\emph{etc.}}
\newcommand{\cat}[1]{`\textit{#1}'}

\newcommand{\xhdr}[1]{\vspace{1mm} \noindent{{\bf #1.}}}

\newcommand{\bfast}{Breakfast}
\newcommand{\mult}{Multi-THUMOS}
\newcommand{\act}{Activity-Net}

\theoremstyle{definition}
\newtheorem{definition}{Definition}%[section]
\newtheorem*{problem*}{Problem Statement}%[section]

\renewcommand{\comment}[1]{}
\usepackage{xcolor}
\usepackage{multirow,amsmath,booktabs,dcolumn}
\usepackage{todonotes}
\usepackage{colortbl}
\usepackage{paralist}
\usepackage{graphicx}
\usepackage{caption}
\usepackage{lipsum}
\usepackage{mathtools}

\SetAlFnt{\small}
\usepackage{xcolor,colortbl}
\newcolumntype{a}{>{\columncolor{blue!5}}c}
\newcolumntype{b}{>{\columncolor{red!5}}c}

\begin{document}
\fancyhead{}

\title{\ourm: Self-Attentive Temporal Point Process Flows \\for Activity Sequences}
\author{Vinayak Gupta}
\affiliation{
  \institution{IIT Delhi}
  \city{New Delhi}
  \country{India}
}
\email{vinayak.gupta@cse.iitd.ac.in}

\author{Srikanta Bedathur}
\affiliation{
  \institution{IIT Delhi}
  \city{New Delhi}
  \country{India}
}
\email{srikanta@cse.iitd.ac.in}

\setlength{\abovedisplayskip}{1pt}
\setlength{\belowdisplayskip}{1pt}

\begin{abstract}
Any human activity can be represented as a temporal sequence of actions performed to achieve a certain goal. Unlike machine-made time series, these action sequences are highly disparate as the time taken to finish a similar action might vary between different persons. Therefore, understanding the dynamics of these sequences is essential for many downstream tasks such as activity length prediction, goal prediction, \etc\ Existing neural approaches that model an activity sequence are either limited to visual data or are task-specific, \ie, limited to next action or goal prediction. In this paper, we present \ourm, a neural marked temporal point process (MTPP) framework for modeling the continuous-time distribution of actions in an activity sequence while simultaneously addressing three high-impact problems -- next action prediction, sequence-goal prediction, and \textit{end-to-end} sequence generation. Specifically, we utilize a self-attention module with temporal normalizing flows to model the influence and the inter-arrival times between actions in a sequence. Moreover, for time-sensitive prediction, we perform an \textit{early} detection of sequence goal via a constrained margin-based optimization procedure. This in-turn allows \ourm to predict the sequence goal using a limited number of actions. Extensive experiments on sequences derived from three activity recognition datasets show the significant accuracy boost of \ourm over the state-of-the-art in terms of action and goal prediction, and the first-ever application of end-to-end action sequence generation.
\end{abstract}

\begin{CCSXML}
<ccs2012>
   <concept>
       <concept_id>10002951.10003227.10003351</concept_id>
       <concept_desc>Information systems~Data mining</concept_desc>
       <concept_significance>500</concept_significance>
       </concept>
 </ccs2012>
\end{CCSXML}
\ccsdesc[500]{Information systems~Data mining}

\keywords{marked temporal point process; continuous time sequences; activity modeling; goal prediction; sequence generation}
\maketitle

\section{Introduction}\label{sec:intro}
A majority of data generated via human activities, \eg, running, playing basketball, cooking, \etc, can be represented as a sequence of actions over a continuous-time. These actions denote a step taken by a user towards achieving a certain goal and vary in their start and completion times, depending on the user and the surrounding environment\cite{mehrasa2017learning,breakfast,avae}. Therefore, unlike synthetic time series, these continuous-time action sequences (CTAS) can vary significantly even if they consist of the same set of actions. For \eg, one person making omelets may take a longer time to cook eggs while another may prefer to cook for a short time\footnote{\url{https://bit.ly/3F5aEwX} (Accessed January 2022)}; or Xavi may make a quicker pass than Pirlo in a football game -- although in both cases the goals are same, and they all have to perform the same sequence of actions. In addition, modeling the dynamics of these CTAS is increasingly challenging due to the limited ability of modern recurrent and self-attention-based approaches in capturing continuous action times~\cite{transformer,sasrec}. This situation is further exacerbated due to the asynchronous occurrence of actions and the large variance in action-\textit{times} and \textit{types}. Therefore, the problem of modeling a CTAS has been overlooked by the past literature.

In recent years, neural marked temporal point processes (MTPP) have shown a significant promise in modeling a variety of continuous-time sequences in healthcare~\cite{rizoiu2,rizoiu1}, finance~\cite{sahp,bacry}, education~\cite{sahebi}, and social networks~\cite{leskovec,nhp,thp}. However, standard MTPP have a limited modeling ability for CTAS as: (i) they assume a homogeneity among sequences, \ie, they cannot distinguish between two sequences of similar actions but with different time duration; (ii) in a CTAS, an action may finish before the start of the next action and thus, to model this empty time interval an MTPP must introduce a new action type, \ie, \textit{NULL} or \textit{end-action} which may lead to an unwarranted increase in the types of actions to be modeled; and (iii) they cannot encapsulate the additional features associated with an action, for \eg, minimum time for completion, necessary previous actions, or can be extended to sequence generation.

\subsection{Our Contribution}
In this work, we present \ourm (\textbf{P}oint P\textbf{ro}cess flows for \textbf{Activ}ity S\textbf{e}qeunces), a normalizing flow-based neural MTPP framework, designed specifically to model the dynamics of a CTAS. Specifically, \ourm addresses three challenging problems -- (i) action prediction; (ii) goal detection; and (iii) the first-of-its-kind task of \textit{end-to-end} sequence generation. We learn the distribution of actions in a CTAS using temporal normalizing flows (NF)~\cite{shakir,ppflows} conditioned on the dynamics of the sequence as well as the action-features (\eg, minimum completion time, \etc). Such a flow-based formulation provides \ourm flexibility over other similar models~\cite{karishma,intfree} to better model the inter-action dynamics within and across sequences. Moreover, our model is designed for \textit{early} detection of the sequence goal, \ie, identifying the result of a CTAS without traversing the complete sequence. We achieve this via a time-bounded optimization, \ie, by incrementally increasing the probability of identifying the true goal using a \textit{margin}-based and a weighted factor-based learning~\cite{margin,earlyijcv}. Such an optimization procedure allows \ourm to model the goal-action hierarchy, \ie, the \textit{necessary} set of actions in a CTAS towards achieving a particular goal, and simultaneously, the order of actions in CTAS with similar goals.

To the best of our knowledge, in this paper we present the first-ever application of MTPP via \textit{end-to-end} action sequence generation. Specifcally, given the resultant goal, \ourm can generate a CTAS with the necessary set of actions and their occurrence times. Such a novel ability for MTPP models can reinforce their usage in applications related to bio-signals~\cite{haradal}, sensor-data~\cite{sensegen}, \etc, and overcome the modeling challenge due to scarcity of activity data~\cite{donahue2020user,luo2020database,axolotl}. Buoyed by the success of attention models in sequential applications~\cite{transformer}, we use a self-attention architecture in \ourm to model the inter-action influences in a CTAS. In summary, the key contributions we make in this paper via \ourm are:
\begin{enumerate}
\item We propose \ourm, a novel temporal flow-based MTPP framework designed specifically for modeling human activities with a time-bounded optimization framework for early detection of CTAS goal. 
\item Our normalizing flow-based modeling framework incorporates the sequence and individual action dynamics along with the action-goal hierarchy. Thus, \ourm introduces the first-of-its-kind MTPP application of end-to-end action CTAS generation with just the sequence-goal as input.
\item Finally, we empirically show that \ourm outperforms the state-of-the-art models for all three tasks --  action prediction, goal detection, and sequence generation.
\end{enumerate}

\subsection{Organization}
We present a problem formulation and a background on necessary techniques in Section~\ref{sec:basics}. Section~\ref{sec:model} gives an overview followed by an detailed development of all components in \ourm. Section~\ref{sec:exp} contains in-depth experimental analysis and qualitative studies over all datasets. Lastly, Section~\ref{sec:related} reviews a few relevant works before concluding in Section~\ref{sec:conc}.

\section{Preliminaries}\label{sec:basics}
In this section, we present a background of MTPP and normalizing flows, and then present the problems addressed in this paper.

\subsection{Background} \label{app:back}
\xhdr{Marked Temporal Point Processes}
MTPP\cite{hawkes} are probabilistic generative models for continuous-time event sequences. An MTPP can be represented as a probability distribution over sequences of variable length belonging to a time interval $[0, T]$. Equivalently, they can be described using a counting process, say $N(t)$, and are characterized by the underlying conditional intensity function, $\lambda^*(t)$ which specifies the likelihood of the next event, conditioned on the history of events. The intensity function $\lambda^*(t)$ computes the infinitesimal probability that an event will happen in the time window $(t, t + dt]$ conditioned on the history as:
\begin{equation}
\mathbb{P} \big(d N(t) = N(t+dt) - N(t) = 1 \big) = \lambda^*(t),
\end{equation}
Here, $*$ denotes a dependence on history. Given the conditional intensity function, we obtain the probability density function as:
\begin{equation}
p^*(\Delta_{t, i}) = \lambda^*(t_{i-1} + \Delta_{t, i}) \exp \bigg(-\int_{0}^{\Delta_{t, i}} \lambda^*(t_{i-1} + r) dr\bigg),
\end{equation}
where, $\Delta_{t, i}$ denotes the inter-event time interval, \ie, $t_i - t_{i-1}$. In contrast to other neural MTPP models that rely on the intensity function~\cite{rmtpp,nhp,sahp,thp} we replace the intensity function with a \emph{log-normal} flow. Such a formulation facilitates closed-form and faster sampling as well as more accurate prediction than the intensity-based models~\cite{intfree,ppflows}.

\xhdr{Normalizing Flows}
Normalizing flows\cite{shakir,intfree,ppflows} are generative models that are used for density estimation as well as event sampling. They work by mapping simple distributions to complex ones using multiple bijective functions. In detail, if our goal is to estimate the density function $p_{\bs{X}}$ of a random vector $\bs{X} \in \mathbb{R}^D$, then the normalizing flows assign a new distribution $\bs{X} = g_{\phi}(\bs{Z})$, with $g_{\phi}: \mathbb{R}^D \rightarrow \mathbb{R}^D$ is a bijective function, and $\bs{Z} \in \mathbb{R}^D$ is a random vector samped from a simple density function $p_{\bs{Z}}$. Using NFs,  sampling an event from $p_{\bs{X}}$ is done by a two step procedure of first sampling from the simple distribution $\bs{z} \sim p_{\bs{Z}}$ and then applying the bijective function $\bs{x} = g_{\phi}(\bs{z})$. Such a procedure supports closed form sampling. Moreover, modern approaches\cite{autoregressive,dhaliwal,shakir} represent the function $g_{\phi}(\bullet)$ via a neural network.

\subsection{Problem Formulation}
As mentioned in Section~\ref{sec:intro}, we represent an activity via a continuous-time action sequence, \ie, a series of actions undertaken by users and their corresponding time of occurrences. We derive each CTAS from annotated frames of videos consisting of individuals performing certain activities. Specifically, for every video, we have a sequence of activity labels being performed in the video along with timestamps for each activity. Therefore, each CTAS used in our dataset is derived from these sequences of a video. Formally, we provide a detailed description of a CTAS in Definition~\ref{def: ctas}.
\begin{definition}[Continuous Time Action Sequence]
\label{def: ctas}
\textit{We define a continuous-time action sequence (CTAS) as a series of action events taken by a user to achieve a particular goal. Specifically, we represent a CTAS as $\cm{S}_k=\{e_i=(c_i, t_i) | i \in[k] , t_i<t_{i+1}\}$, where $t_i \in \mathbb{R}^+$ is the start-time of the action, $c_i\in \cm{C}$ is the discrete category or mark of the $i$-th action, $\cm{C}$ is the set of all categories, $\Delta_{t, i} = t_i - t_{i-1}$ as the inter-action time, and $\cm{S}_k$ denotes the sequence of first $k$ actions. Each CTAS has an associated result, $g \in \cm{G}$, that signifies the goal of the CTAS. Here, $\cm{G}$ denotes the set of all possible sequence goals.}
\end{definition}

To highlight the relationship between sequence goal and actions consider the example of a CTAS with the goal of \cat{making-coffee}, would comprise of actions -- \cat{take-a-cup}, \cat{pour-milk}, \cat{add-coffee-powder}, \cat{add-sugar}, and \cat{stir} -- at different time intervals. Given the aforementioned definitions, we formulate the tasks of next action, sequence goal prediction, and sequence generation as:

\xhdr{Input}
A CTAS of all actions, $\cm{S}_k$, consisting of categories and times of different actions that lead to a goal $g$.

\xhdr{Output}
A probabilistic prediction model with three distinct tasks -- (i) to estimate the likelihood of the next action $e_{k+1}$ along with the action category and occurrence time; (ii) to predict the goal of the CTAS being modeled, \ie, $\widehat{g}$; and (iii) a generative model to sample a sequence of actions, $\widehat{\cm{S}}$ given the true sequence goal, $g$.
\section{\ourm Model}\label{sec:model}
In this section, we first present a high-level overview of the \ourm model and then describe the neural parameterization of each component in detail. Lastly, we provide a detailed description of its optimization and sequence generation procedure.

\subsection{High Level Overview}\label{sec:overview}
We use an MTPP denoted by $p_{\theta}(\cdot)$, to learn the generative mechanism of a continuous-time action sequence. Moreover, we design the sequence modeling framework of $p_{\theta}(\cdot)$ using a self-attention based encoder-decoder model~\cite{transformer}. Specifically, we embed the actions in a CTAS, \ie, $\cm{S}_k$, to a vector embedding, denoted by $\bs{s}_k$, using a weighted aggregation of all past actions. Therefore, $\bs{s}_k$ signifies a compact neural representation of the sequence history, \ie, all actions till the $k$-th index and their marks and occurrence times. Recent research~\cite{thp,sahp} has shown that an attention-based modeling choice can better capture the long-term dependencies as compared to RNN-based MTPP models~\cite{rmtpp,nhp,intfree,fullyneural}. A detailed description of the embedding procedure is given in Section~\ref{sec:detail}.

We use our MTPP $p_{\theta}(\cdot)$ to estimate the generative model for the $(k+1)$-th action conditioned on the past, \ie, $p(e_{k+1})$ as:
\begin{equation}
p_{\theta}(e_{k+1} | \bs{s}_k) = \mathbb{P}_{\theta}(c_{k+1}|\bs{s}_k) \cdot \rho_{\theta}(\Delta_{t, k+1}|\bs{s}_k),
\end{equation}
where, $\mathbb{P}_{\theta}(\cdot)$ and $\rho_{\theta}(\cdot)$ denote the probability distribution of marks and the density function for inter-action arrival times respectively. Note that both the functions are conditioned on $\bs{s}_k$ and thus \ourm requires a joint optimizing procedure for both -- action time and mark prediction. Next, we describe the mechanism used in \ourm to predict the next action and goal detection in a CTAS.

\xhdr{Next Action Prediction}
We determine the most probable mark and time of the next action, using $p_{\theta} (\cdot)$ via standard sampling techniques over $\mathbb{P}_{\theta}(\cdot)$ and $\rho_{\theta}(\cdot)$ respectively~\cite{rmtpp,intfree}.
\begin{equation}
\widehat{e_{k+1}} \sim p_{\theta}(e_{k+1} | \bs{s}_k),
\end{equation}
In addition, to keep the history embedding up-to-date with the all past actions, we iteratively update $\bs{s}_k$ to $\bs{s}_{k+1}$ by incorporating the details of action $e_{k+1}$.

\xhdr{Goal Detection}
Since the history embedding, $\bs{s}_k$, represents an aggregation of all past actions in a sequence, it can also be used to capture the influences between actions and thus, can be extended to detect the goal of the CTAS. Specifically, to detect the CTAS goal, we use a non-linear transformation over $\bs{s}_k$ as:
\begin{equation}
\widehat{g} \sim \mathbb{P}_{g' \in \cm{G}}(\Phi(s_k)),
\label{eq:goal}
\end{equation}
where, $\mathbb{P}_{\bullet}$ denotes the distribution over all sequence goals and $\Phi(\cdot)$ denotes the transformation via a fully-connected MLP layer.

\subsection{Neural Parametrization}\label{sec:detail}
Here, we present a detailed description of the neural architecture of our MTPP, $p_{\theta}(\cdot)$, and the optimization procedure in \ourm. Specifically, we realize $p_{\theta}(\cdot)$ using a three layer architecture:

\xhdr{Input Layer}
As mentioned in Section~\ref{sec:basics}, each action $e_i \in \cm{S}_k$ is represented by a mark $c_i$ and time $t_i$. Therefore, we embed each action as a combination of all these features as:
\begin{equation}
\bs{y}_i = \bs{w}_{y, c} c_i + \bs{w}_{y, t} t_{i} + \bs{w}_{y, \Delta} \Delta_{t,i} + \bs{b}_y,
\end{equation}
where $\bs{w}_{\bullet, \bullet}, \bs{b}_{\bullet}$ are trainable parameters and $\bs{y}_i \in \mathbb{R}^D$ denotes the vector embedding for the action $e_i$ respectively. In other sections as well, we denote weight and bias as $\bs{w}_{\bullet, \bullet}$ and $\bs{b}_{\bullet, \bullet}$ respectively. 

\xhdr{Self-Attention Layer}
We use a \textit{masked} self-attention layer to embed the past actions to $\bs{s}_k$ and to interpret the influence between the past and the future actions. In detail, we follow the standard attention procedure~\cite{transformer} and first add a trainable positional encoding, $\bs{p}_i$, to the action embedding, \ie, $\bs{y}_i \leftarrow \bs{y}_i + \bs{p}_i$. Such trainable encodings are shown to be more scalable and robust for long sequence lengths as compared to those based on a fixed function~\cite{sasrec,tisasrec}. Later, to calculate an attentive aggregation of all actions in the past, we perform three independent linear transformations on the action representation to get the \textit{query}, \textit{key}, and \textit{value} embeddings, \ie, 
\begin{equation}
\bs{q}_i = \bs{W}^Q  \bs{y}_i, \quad \bs{k}_i = \bs{W}^K  \bs{y}_i, \quad \bs{v}_i = \bs{W}^V  \bs{y}_i,
\end{equation}
where, $\bs{q}_{\bullet}, \bs{k}_{\bullet}, \bs{v}_{\bullet}$ denote the query, key, and value vectors respectively. Following standard self-attention model, we represent $\bs{W}^{Q}$, $\bs{W}^{K}$ and $\bs{W}^{V}$ as trainable \textit{Query}, \textit{Key}, and \textit{Value} matrices respectively. Finally, we compute $\bs{s}_k$ conditioned on the history as:
\begin{equation}
\bs{s}_k =  \sum_{i=1}^{k} \frac{\exp\left(\bs{q}_k^{\top} \bs{k}_i /\sqrt{D} \right)}{\sum_{i'=1}^{k}\exp\left( \bs{q}_k^{\top} \bs{k}_{i'} /\sqrt{D} \right)} \bs{v}_i, \label{eq:attn}
\end{equation}
where, $D$ denotes the number of hidden dimensions. Here, we compute the attention weights via a softmax over the interactions between the query and key embeddings of each action in the sequence and perform a weighted sum of the value embeddings.

Now, given the representation $\bs{s}_k$, we use the attention mechanism in Eqn.~\eqref{eq:attn} and apply a feed-forward neural network to incorporate the necessary non-linearity to the model as:
\begin{equation*}
\bs{s}_k \leftarrow \sum_{i=1}^k \big[ \bs{w}_{s, m}\odot\textsc{ReLU}(\bs{s}_i \odot \bs{w}_{s, n} + \bs{b}_{s, n})  + \bs{b}_{s, m} \big],
\end{equation*}
where, $\bs{w}_{s, m}, \bs{b}_{s, m}$ and $\bs{w}_{s, n}, \bs{b}_{s, n}$ are trainable parameters of the outer and inner layer of the point-wise feed-forward layer.  

To support faster convergence and training stability, we employ: (i) layer normalization; (ii) stacking multiple self-attention blocks; and (iii) multi-head attention. Since these are standard techniques~\cite{ba2016layer,transformer}, we omit their mathematical descriptions in this paper.

\xhdr{Output Layer}
At every index $k$, \ourm outputs the next action and the most probable goal of the CTAS. We present the prediction procedure for each of them as follows:

\noindent \textit{\underline{Action Prediction:}}
We use the output of the self-attention layer, $\bs{s}_k$ to estimate the mark distribution and time density of the next event, \ie, $\mathbb{P}_{\theta}(e_{k+1})$ and $\rho_{\theta}(e_{k+1})$ respectively. Specifically, we model the $\mathbb{P}_{\theta}(\cdot)$ as a softmax over all other marks as:
\begin{equation}
\mathbb{P}_{\theta}(c_{k+1}) = \frac{\exp\left(\bs{w}_{c, s}^{\top} \bs{s}_i + \bs{b}_{c, s} \right)}{\sum_{c'=1}^{|\cm{C}|}\exp\left( \bs{w}_{c', s}^{\top} \bs{s}_i + \bs{b}_{c', s} \right)},
\label{eqn: samplemark}
\end{equation}
where, $\bs{w}_{\bullet, \bullet}$ and $\bs{b}_{\bullet, \bullet}$ are trainable parameters.

In contrast to standard MTPP approaches that rely on an intensity-based model~\cite{rmtpp,nhp,thp,sahp}, we capture the inter-action arrival times via a \textit{temporal} normalizing flow (NF). In detail, we use a \textit{LogNormal} flow to model the temporal density $\rho_{\theta}(\Delta_{t, k+1})$. Moreover, standard flow-based approaches~\cite{intfree,ppflows} utilize a common NF for all events in a sequence, \ie, the arrival times of each event are determined from a single or mixture of flows trained on all sequences. We highlight that such an assumption restricts the ability to model the dynamics of a CTAS, as unlike standard events, an action has three distinguishable characteristics -- (i) every action requires a minimum time for completion; (ii) the time taken by a user to complete an action would be similar to the times of another user; and (iii) similar actions require similar times to complete. For example, the time taken to complete the action \cat{add-coffee} would require a certain minimum time of completion and these times would be similar for all users. Intuitively, the time for completing the action \cat{add-coffee} would be similar to those for the action \cat{add-sugar}. 

To incorporate these features in \ourm, we identify actions with similar completion times and model them via independent temporal flows. Specifically, we cluster all actions $c_i \in \cm{C}$ into $\cm{M}$ non-overlapping clusters based on the \textit{mean} of their times of completion and for each cluster we define a trainable embedding $\bs{z}_r \in \mathbb{R}^{D} \, \forall r \in \{1, \cdots, \cm{M}\}$. Later, we sample the start-time of the future action by conditioning our temporal flows on the cluster of the current action as:
\begin{equation}
\widehat{\Delta_{t, k+1}} \sim \textsc{LogNormal}\left(\bs{\mu}_k , \bs{\sigma}^2_k \right),
\label{eqn: sampletime}
\end{equation}
where, $[\bs{\mu}_k ,\bs{\sigma}^2_k]$, denote the mean and variance of the log-normal temporal flow and are calculated via the sequence embedding and the cluster embedding as:
\begin{equation}
\bs{\mu}_k = \sum_{r=1}^{\cm{M}} \cm{R}(e_k, r) \big(\bs{w}_{\mu} \left(\bs{s}_{k}\odot\bs{z}_{c, i} \right) + \bs{b}_{\mu}\big),
\end{equation}
\begin{equation}
\bs{\sigma}^2_k  = \sum_{r=1}^{\cm{M}} \cm{R}(e_k, r)  \big( \bs{w}_{\sigma} \left(\bs{s}_{k}\odot\bs{z}_{c, i} \right) + \bs{b}_{\sigma}\big),
\end{equation}
where $\bs{w}_{\bullet}, \bs{b}_{\bullet}$ are trainable parameters, $\cm{R}(e_k, r)$ is an indicator function that determines if event $e_k$ belongs to the cluster $r$ and $\bs{z}_r$ denotes the corresponding cluster embedding. Such a cluster-based formulation facilitates the ability of model to assign similar completion times for events in a same cluster. To calculate the time of next action, we add the sampled time difference to the time of the previous action $e_k$, \ie,
\begin{equation}
\widehat{t_{k+1}} = t_k + \widehat{\Delta_{t, k+1}}
\end{equation}
where, $\widehat{t_{k+1}}$ denotes the predicted time for the action $e_{k+1}$. 

\noindent \textit{\underline{Goal Detection:}}
In contrast to other MTPP approaches~\cite{rmtpp,nhp,thp,sahp,intfree}, an important feature of \ourm is identifying the goal of a sequence, \ie, a hierarchy on top of the actions in a sequence, based on the past sequence dynamics. To determine the goal of a CTAS, we utilize the history embedding $\bs{s}_k$ as it encodes the inter-action relationships of all actions in the past. Specifically, we use a non-linear transformation via a feed-forward network, denoted as $\Phi(\cdot)$ over $\bs{s}_k$ and apply a softmax over all possible goals.
\begin{equation}
\Phi(\bs{s}_k) = \textsc{ReLU} (\bs{w}_{\Phi, s} \bs{s}_k + \bs{b}_{\Phi, s}),
\end{equation}
where, $\bs{w}_{\bullet, \bullet}, \bs{b}_{\bullet, \bullet}$ are trainable parameters. We sample the most probable goal as in Eqn.~\eqref{eq:goal}. 

We highlight that we predict the CTAS goal at each interval, though a CTAS has only one goal. This is to facilitate \textit{early} goal detection in comparison to detecting the goal after traversing the entire CTAS. More details are given in Section~\ref{sec:early} and Section~\ref{sec:optimization}.

\subsection{Early Goal Detection and Action Hierarchy}\label{sec:early}
Here, we highlight the two salient features of \ourm -- early goal detection and modeling the goal-action hierarchy.

\xhdr{Early Goal Detection}
Early detection of sequence goals has many applications ranging from robotics to vision~\cite{earlyiccv,earlyijcv}. To facilitate early detection of the goal of a CTAS in \ourm, we devise a ranking loss that forces the model to predict a \textit{non-decreasing} detection score for the correct goal category. Specifically, the detection score of the correct goal at the $k$-th index of the sequence, denoted by $p_k(g| \bs{s}_k, \Phi)$, must be more than the scores assigned the correct goal in the past. Formally, we define the ranking loss as:
\begin{equation}
\cm{L}_{k, g} = \max \big(0, p^*_k(g) - p_k(g| \bs{s}_k, \Phi)\big),
\label{eqn: margin}
\end{equation}
where $p^*_k(g)$ denotes the maximum probability score given to the correct goal in all past predictions.
\begin{equation}
p^*_k(g) = \max_{j \in \{1, k-1\}} p_j(g| \bs{s}_j, \Phi),
\label{eqn: pastmax}
\end{equation}
where $p_k(g)$ denotes the probability score for the correct goal at index $k$. Intuitively, the ranking loss $\cm{L}_{k,g}$ would penalize the model for predicting a smaller detection score for the correct CTAS goal than any previous detection score for the same goal.

\xhdr{Action Hierarchy}
Standard MTPP approaches assume the category of marks as independent discrete variables, \ie, the probability of an upcoming mark is calculated independently~\cite{rmtpp,nhp,sahp,thp}. Such an assumption restricts the predictive ability while modeling CTAS, as in the latter case, there exists a hierarchy between goals and actions that lead to the specific goal. Specifically, actions that lead to a common goal may have similar dynamics and it is also essential to model the relationships between the actions of different CTAS with a common goal. We incorporate this hierarchy in \ourm along with our next action prediction via an action-based ranking loss. In detail, we devise a loss function similar to Eqn.~\ref{eqn: margin} where we restrict the model to assign non-decreasing probabilities to all actions leading to the goal of CTAS under scrutiny. 
\begin{equation}
\cm{L}_{k, c} = \sum_{c' \in \cm{C}^*_g} \max \big(0, p^*_k(c') - p_k(c'| \bs{s}_k)\big),
\label{eqn: cat_margin}
\end{equation}
where $\cm{C}^*_g, p_k(c'| \bs{s}_k)$ denote a set of all actions in CTAS with the goal $g$ and the probability score for the action $c' \in \cm{C}^*_g$ at index $k$ respectively. Here, $p^*_k(c')$ denotes the maximum probability score given to action $c'$ in all past predictions and is calculated similar to Eqn.~\ref{eqn: pastmax}. We regard  $\cm{L}_{k, g}$ and $\cm{L}_{k, c}$ as \textit{margin} losses, as they aim to increase the difference between two prediction probabilities.

\subsection{Optimization}\label{sec:optimization}
We optimize the trainable parameters in \ourm, \ie, the weight and bias tensors ($\bs{w}_{\bullet, \bullet}$ and $\bs{b}_{\bullet, \bullet}$) for our MTPP $p_{\theta} (\cdot)$, using a two channels of training consisting of action and goal prediction. Specifically, to optimize the ability of \ourm for predicting the next action, we maximize the the joint likelihood for the next action and the log-normal density distribution of the temporal flows.
\begin{equation}
\mathscr{L} = \sum_{k = 1}^{|\cm{S}|} \log \big( \mathbb{P}_{\theta}(c_{k+1}|\bs{s}_k) \cdot \rho_{\theta} (\Delta_{t, k+1}| \bs{s}_k) \big ),
\label{eqn:likelihood}
\end{equation}
where $\mathscr{L}$ denotes the joint likelihood, which we represent as the sum of the likelihoods for all CTAS. In addition to action prediction, we optimize the \ourm parameters for \textit{early} goal detection via a temporally weighted cross entropy (CE) loss over all sequence goals. Specifically, we follow a popular reinforcement recipe of using a time-varying \textit{discount} factor over the prediction loss as:
\begin{equation}
\cm{L}_g = \sum_{k = 1}^{|\cm{S}|} \gamma^k \cdot \cm{L}_{\textsc{CE}} \big(p_k(g| \bs{s}_k)\big),
\label{eqn: discount}
\end{equation}
where $\gamma \in [0,1], \cm{L}_{\textsc{CE}} \big(p_k(g| \bs{s}_k)\big)$ denote the decaying factor and a standard softmax-cross-entropy loss respectively. Such a recipe is used exhaustively for faster convergence of reinforcement learning models~\cite{rl_book,discount}. Here, the discount factor penalizes the model for taking longer times for detecting the CTAS goal by decreasing the gradient updates to the loss.

\xhdr{Margin Loss}
In addition, we minimize the margin losses given in Section~\ref{sec:early} with the current optimization procedure. Specifically, we minimize the following loss:
\begin{equation}
\cm{L}_m = \sum_{k=1}^{|\cm{S}|} \cm{L}_{k, g} + \cm{L}_{k, c},
\end{equation}
where $\cm{L}_{k, g}$ and $\cm{L}_{k, c}$ are margin losses defined in Eqn.~\ref{eqn: margin} and Eqn.~\ref{eqn: cat_margin} respectively. We learn the parameters of \ourm using an Adam~\cite{adam} optimizer for both likelihood and prediction losses.

\subsection{Sequence Generation}\label{sec:generation}
A crucial contribution of this paper via \ourm is an end-to-end generation of action sequences. Specifically, given the CTAS goal as input, we can generate a most probable sequence of actions that may lead to that specific goal. Such a feature has a range of applications from sports analytics~\cite{mehrasa2017learning}, forecasting~\cite{prathamesh}, identifying the duration of an activity~\cite{avae}, \etc

A standard approach for training a sequence generator is to sample future actions in a sequence and then compare with the true actions~\cite{timegan}. However, such a procedure has multiple drawbacks as it is susceptible to noises during training and deteriorates the scalability of the model. Moreover, we highlight that such a sampling based training cannot be applied to a self-attention-based model as it requires a fixed sized sequence as input~\cite{transformer}. Therefore, we resort to a two-step generation procedure that is defined below:
\begin{enumerate}
\item \textbf{Pre-Training:} The first step requires training all \ourm parameters for action prediction and goal detection. This step is necessary to model the relationships between actions and goals and we represent the set of optimized parameters as $\theta^*$ and the corresponding MTPP as $p_{\theta^*}(\cdot)$ respectively.
\item \textbf{Iterative Sampling:} We iteratively sample events and update parameters via our trained MTPP till the model predicts the correct goal for CTAS or we encounter an \texttt{<EOS>} action. Specifically, using $p_{\theta^*}(\cdot)$ and the first \textit{real} action ($e_1$) as input, we calculate the detection score for the correct goal, \ie, $p_1(g| \bs{s}_k)$ and while its value is highest among all probable goals, we sample the mark and time of next action using Eqn.~\ref{eqn: samplemark} and Eqn.~\ref{eqn: sampletime} respectively.
\end{enumerate}
Such a generation procedure harnesses the fast sampling of temporal normalizing flows and simultaneously is conditioned on the action and goal relationships. A detailed pseudo-code of sequence generation procedure used in \ourm is given in Algorithm~\ref{axoalgo}.

\begin{algorithm}[t!]
\textbf{Input:} $g$: Goal of CTAS \\ $e_1$: First Action \\ $p_{\theta^*}(\cdot)$: Trained MTPP\\
\textbf{Output:} $\widehat{S}$: Generated CTAS
$\cm{S}_1 \leftarrow e_1$ \\
$k = 1$\\ 
  \While {$k < \mathtt{max\_len}$}
  {
  	Sample the mark of next action: $\widehat{c_{k+1}} \sim \mathbb{P}_{\theta^*}(\bs{s}_k)$\\
  	Sample the time of next action: $\widehat{t_{k+1}} \sim \rho_{\theta^*}(\bs{s}_k)$\\
  	Add to CTAS: $\cm{S}_{k+1} \leftarrow \cm{S}_{k} + e_{k+1}$\\
    Update the MTPP parameters $\bs{s}_{k+1} \leftarrow p(\bs{s}_{k}, e_{k+1})$\\
    Calculate most probable goal: $\widehat{g}_k = \max_{\forall g'} \big(p_k(g'| \bs{s}_k)\big)$\\
    \uIf{$\widehat{g_i} != g \, \mathrm{or} \, \widehat{c_{k+1}} == \mathtt{<EOS>}$}
	{
	Add EOS mark: $\widehat{\cm{S}} \leftarrow \cm{S}_{k+1} + \mathtt{<EOS>}$\\
	Exit the sampling procedure: $\textsc{Break}$\\
	}
	Increment iteration: $k \leftarrow k + 1$\\
	
  }\label{endfor}
  Return generated CTAS: return $\widehat{\cm{S}}$\\
\caption{Sequence Generation with \ourm}\label{axoalgo}
\end{algorithm}

%\subsection{Salient Features}\label{sec:features}
%Here we reiterate the salient features of \ourm in context of the model description in Section~\ref{sec:model}. Specifically, our contribution via \ourm is 
\section{Experiments}\label{sec:exp}
In this section, we present the experimental setup and the empirical results to validate the efficacy of \ourm. Through our experiments we aim to answer the following research questions: 
\begin{itemize}
\item[\textbf{RQ1}] What is the action-mark and time prediction performance of \ourm in comparison to the state-of-the-art baselines?
\item[\textbf{RQ2}] How accurately and quickly can \ourm identify the goal of an activity sequence?
\item[\textbf{RQ3}] How effectively can \ourm generate an action sequence?
\item[\textbf{RQ4}] How does the action prediction performance  of \ourm vary with different hyperparameters values?
\end{itemize}

\begin{table*}[t!]
\small
\caption{Performance of all the methods in terms of action prediction accuracy (APA) and mean absolute error (MAE) across all datasets. Bold fonts and underline indicate the best  performer and the best baseline respectively. Results marked \textsuperscript{$\dagger$} are statistically significant (\ie, two-sided $t$-test with $p \le 0.1$) over the best baseline.}
\vspace{-0.3cm}
\centering
\begin{tabular}{l|ccc|ccc}
\toprule
& \multicolumn{3}{c|}{\textbf{Action Prediction Accuracy (APA)}} & \multicolumn{3}{c}{\textbf{Mean Absolute Error (MAE)}} \\ \hline 
& \bfast & \mult & \act & \bfast & \mult & \act \\ \hline
NHP~\cite{nhp} & 0.528$\pm$0.024 & 0.272$\pm$0.019 & 0.684$\pm$0.034 & 0.411$\pm$0.019 & \underline{0.017$\pm$0.002} & 0.796$\pm$0.045 \\
AVAE~\cite{avae} & 0.533$\pm$0.028 & 0.279$\pm$0.022 & 0.678$\pm$0.036 & 0.417$\pm$0.021 & 0.018$\pm$0.002 & 0.803$\pm$0.049 \\
RMTPP~\cite{rmtpp} & 0.542$\pm$0.022 & 0.274$\pm$0.017 & 0.683$\pm$0.034 & \underline{0.403$\pm$0.018} & \underline{0.017$\pm$0.002} & \underline{0.791$\pm$0.046} \\
SAHP~\cite{sahp} & 0.547$\pm$0.031 & 0.287$\pm$0.023 & 0.688$\pm$0.042 & 0.425$\pm$0.031 & 0.019$\pm$0.003 & 0.820$\pm$0.072 \\
THP~\cite{thp} & \underline{0.559$\pm$0.028} & \underline{0.305$\pm$0.018} & \underline{0.693$\pm$0.038} & 0.413$\pm$0.023 & 0.019$\pm$0.002 & 0.806$\pm$0.061 \\ \hline
\ours-c & 0.561$\pm$0.027 & 0.297$\pm$0.020 & 0.698$\pm$0.038 & 0.415$\pm$0.027 & 0.015$\pm$0.002 & 0.774$\pm$0.054 \\
\ours-t & 0.579$\pm$0.025 & 0.306$\pm$0.018 & 0.722$\pm$0.035 & 0.407$\pm$0.025 & 0.015$\pm$0.002 & 0.783$\pm$0.058 \\
\ourm & \textbf{0.583$\pm$0.027}\textsuperscript{$\dagger$} & \textbf{0.316$\pm$0.019} & \textbf{0.728$\pm$0.037}\textsuperscript{$\dagger$} & \textbf{0.364$\pm$0.028}\textsuperscript{$\dagger$} & \textbf{0.013$\pm$0.002}\textsuperscript{$\dagger$} & \textbf{0.742$\pm$0.059}\textsuperscript{$\dagger$} \\
\bottomrule
\end{tabular}
\label{tab:main}
\end{table*}

\subsection{Datasets}
To evaluate \ourm, we need time-stamped action sequences and their goals. Therefore, we derive CTAS from three activity modeling datasets sourced from different real-world applications -- cooking, sports, and collective activity. The datasets vary significantly in terms of origin, sparsity, and sequence lengths. We highlight the details of each of these datasets below: 
\begin{itemize}
\item \textbf{\bfast~\cite{breakfast}.} This dataset contains CTAS derived from 1712 videos of different people preparing breakfast. The actions in a CTAS and sequence goals can be classified into 48 and 9 classes respectively. These actions are performed by 52 different individuals in 18 different kitchens.

\item \textbf{\mult~\cite{thumos}.} A sports activity dataset that is designed for action recognition in videos.  We derive the CTAS using 400 videos of individuals involved in different sports such as discus throw, baseball, \etc\ The actions and goals can be classified into 65 and 9 classes respectively and on average, there are 10.5 action class labels per video.

\item \textbf{\act~\cite{activitynet}.} This dataset comprises of activity categories collected from 591 YouTube videos with a total of 49 action labels and 14 goals.
\end{itemize}
We highlight that in \act, many of the videos are shot by amateurs in many uncontrolled environments, the variances within the CTAS of the same goal are often large, and the lengths of CTAS vary and are often long and complex.

\subsection{Baselines}
We compare the action prediction performance of \ourm with the following state-of-the-art methods: 
\begin{asparaitem}[]
\item \textbf{RMTPP~\cite{rmtpp}}: A recurrent neural network that models time differences to learn a representation of the past events.
\item \textbf{NHP~\cite{nhp}}: Models an MTPP using continuous-time LSTMs for capturing the temporal evolution of sequences.
\item \textbf{AVAE~\cite{avae}}: A variational auto-encoder based MTPP framework designed specifically for activities in a sequence.
\item \textbf{SAHP~\cite{sahp}}: A self-attention model to learn the temporal dynamics using an aggregation of historical events. 
\item \textbf{THP~\cite{thp}}: Extends the transformer model~\cite{transformer} to include the \textit{conditional} intensity of event arrival and the inter-mark influences.
\end{asparaitem}
We omit comparison with other continuous-time models~\cite{fullyneural,intfree,wgantpp,xiao,hawkes} as they have already been outperformed by these approaches. 

\subsection{Evaluation Criteria}
Given the dataset $\cm{D}$ of $N$ action sequences, we split them into training and test set based on the goal of the sequence. Specifically, for each goal $g \in \cm{G}$, we consider 80\%  of the sequences as the training set and the other last 20\% as the test set. We evaluate \ourm and all baselines on the test set in terms of (i) mean absolute error (MAE) of predicted times of action, and (ii) action prediction accuracy (APA) described as:
\begin{equation}
\mathrm{MAE} = \frac{1}{|\cm{S}|}\sum_{e_i\in \cm{S}}[|t_i-\widehat{t}_i|], \quad \mathrm{APA} = \frac{1}{|\cm{S}|}\sum_{e_i\in \cm{S}} \#(c_i=\widehat{c}_i),
\end{equation}
where, $\widehat{t_i}$ and $\widehat{c_i}$ are the predicted time and type the $i$-th action in test set. Moreover, we follow a similar protocol to evaluate the sequence generation ability of \ourm and other models. For goal prediction, we report the results in terms of accuracy (ratio) calculated across all sequences. We calculate confidence intervals across 5 independent runs.

\subsection{Experimental Setup}
All our implementations and datasets are publicly available at: \texttt{https://github.com/data-iitd/proactive/}.

\xhdr{System Configuration}
All our experiments were done on a server running Ubuntu 16.04. CPU: Intel(R) Xeon(R) Gold 5118 CPU @ 2.30GHz , RAM: 125GB and GPU: NVIDIA Tesla T4 16GB DDR6. 

\xhdr{Parameter Settings}
For our experiments, we set $D=16$, $\cm{M}=8$, $\gamma=0.9$ and weigh the margin loss $\cm{L}_m$ by 0.1. In addition, we set a $l_2$ regularizer over the parameters with coefficient value $0.001$.

\begin{figure}[b]
\centering
\begin{subfigure}[b]{0.46\columnwidth}
\includegraphics[width=\linewidth]{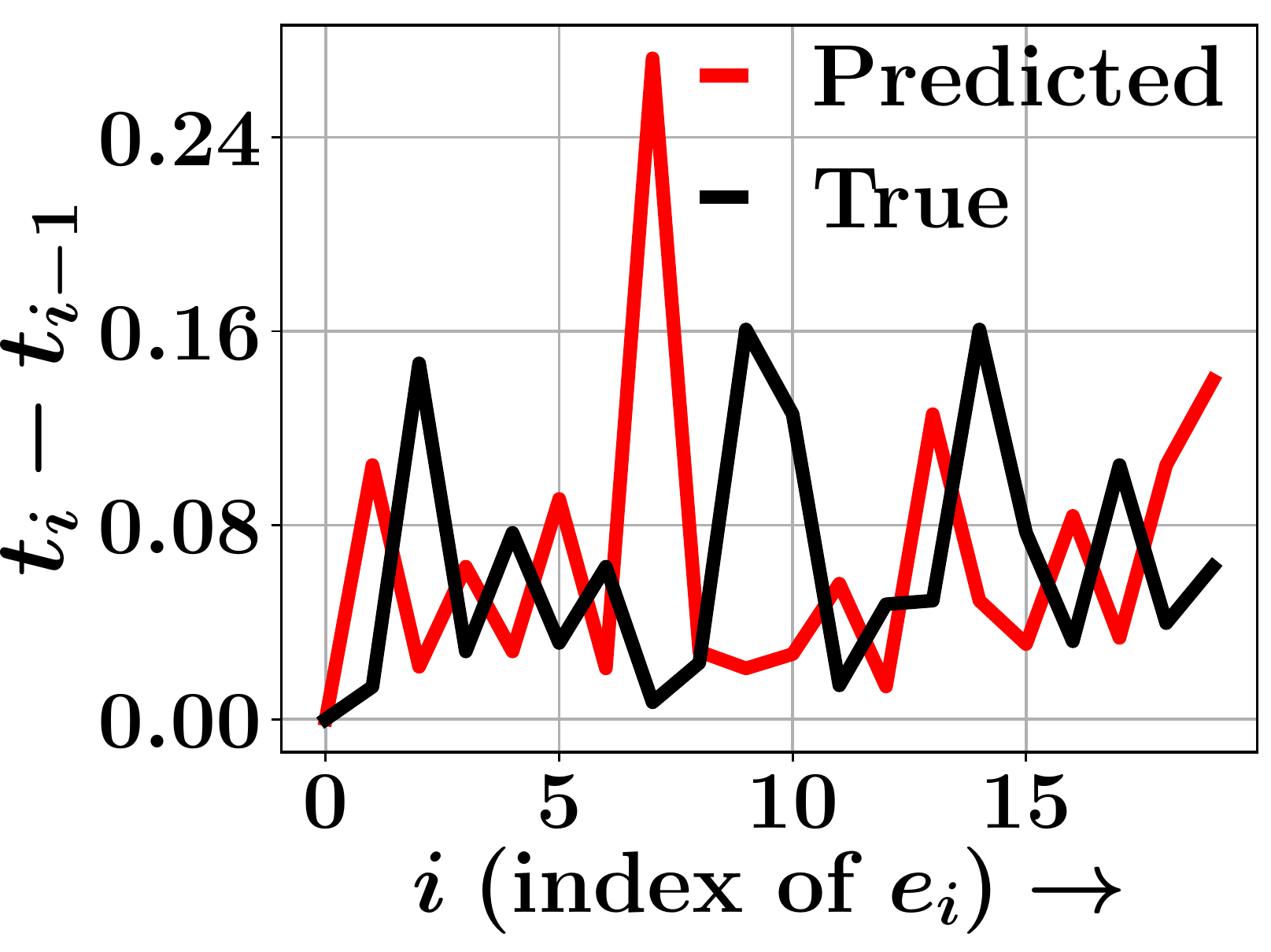}
\caption{\mult}
\end{subfigure}
\hfill
\begin{subfigure}[b]{0.46\columnwidth}
\includegraphics[width=\linewidth]{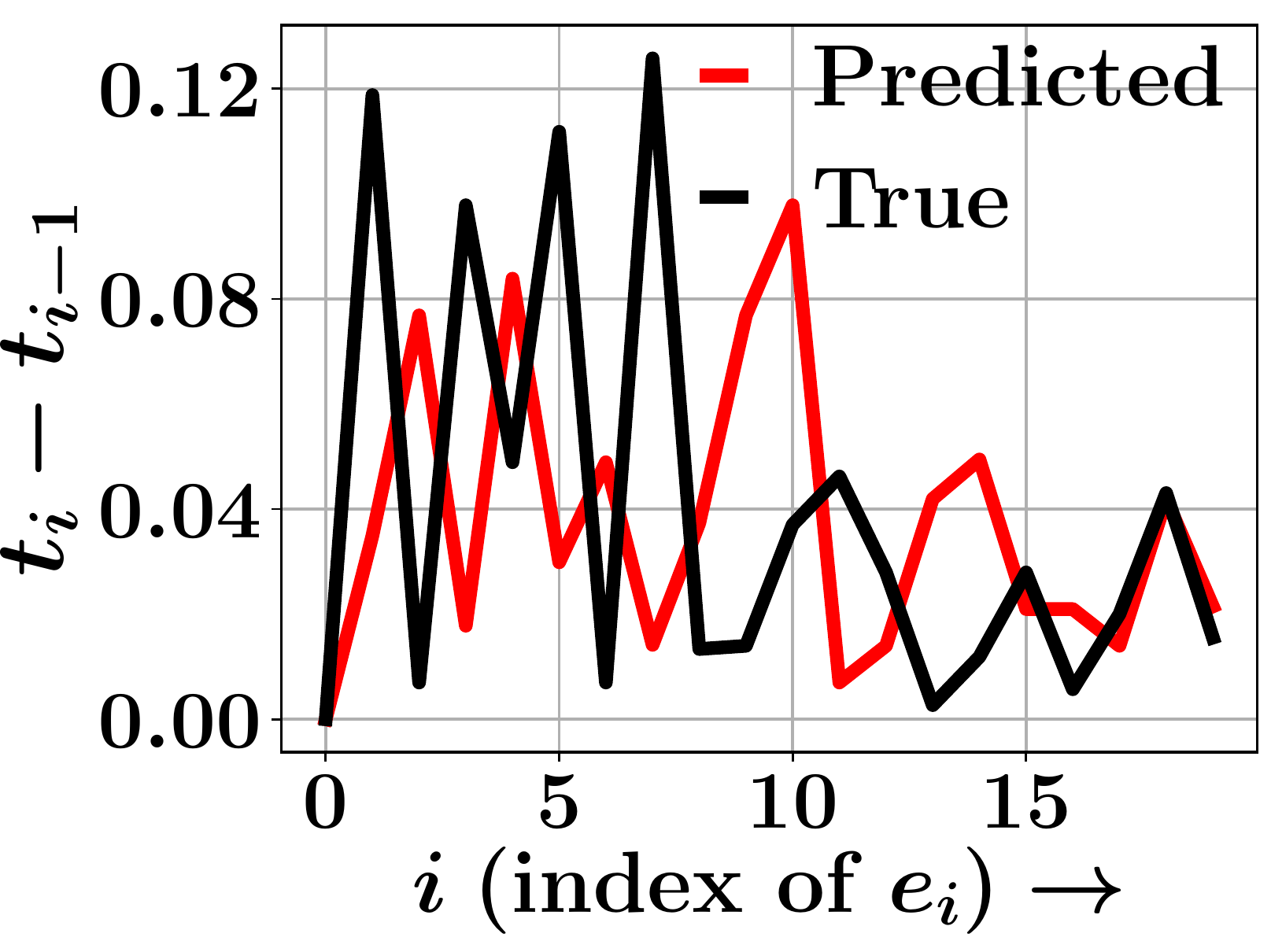}
\caption{\act}
\end{subfigure}
\vspace{-0.3cm}
\caption{\label{fig:qualitative} Real life \textit{true} and \textit{predicted} inter-arrival times $\Delta_{t,k}$ of different events $e_k$ for (a) \mult\ and (b) \act\ datasets. The results show that the true arrival times match with the times predicted by \ourm.}
\end{figure}

\begin{figure*}[t]
\centering
\begin{subfigure}[b]{0.32\linewidth}
{\includegraphics[height=3.2cm]{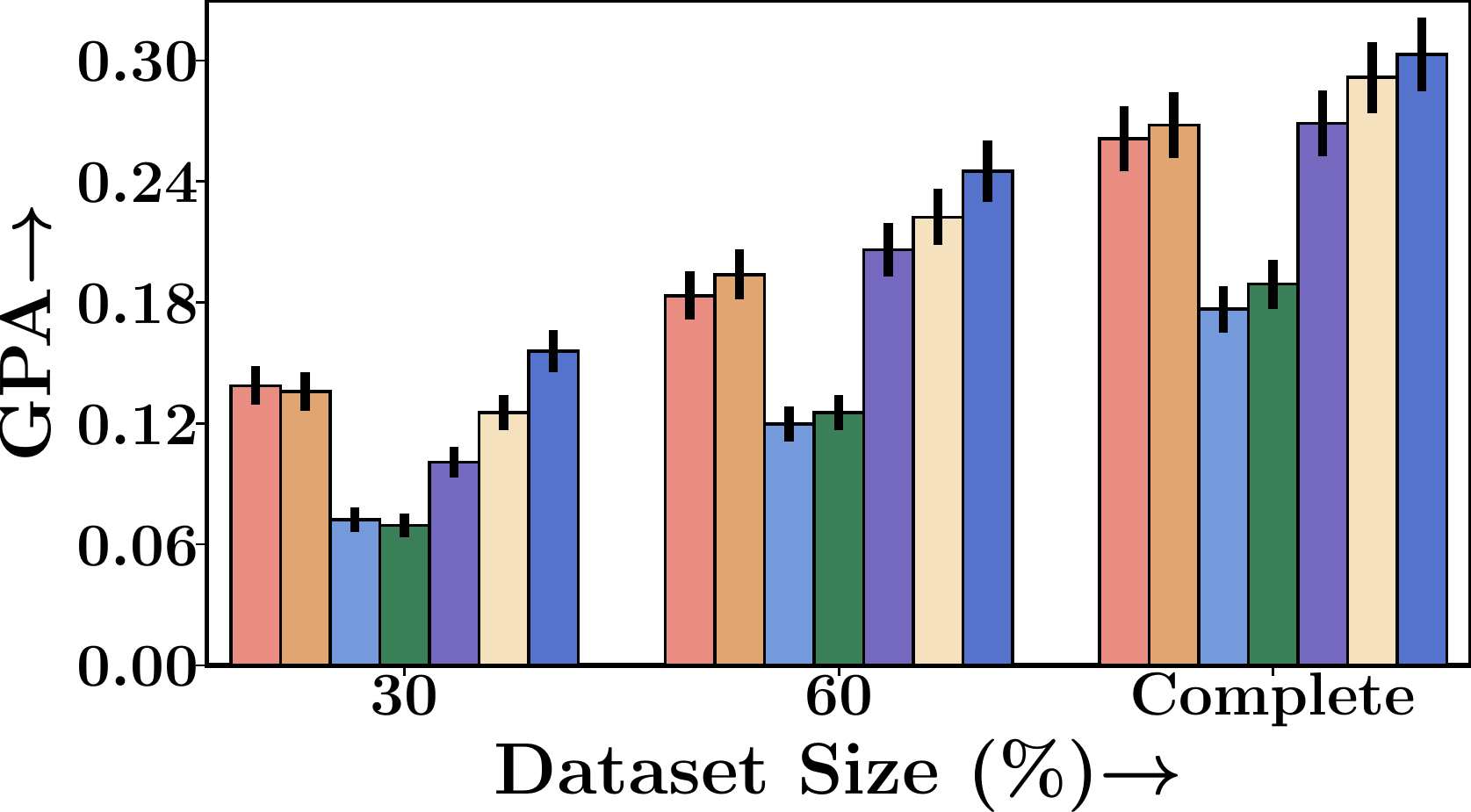}}
\caption{\bfast}
\end{subfigure}
\hfill
\begin{subfigure}[b]{0.32\linewidth}
{\includegraphics[height=3.2cm]{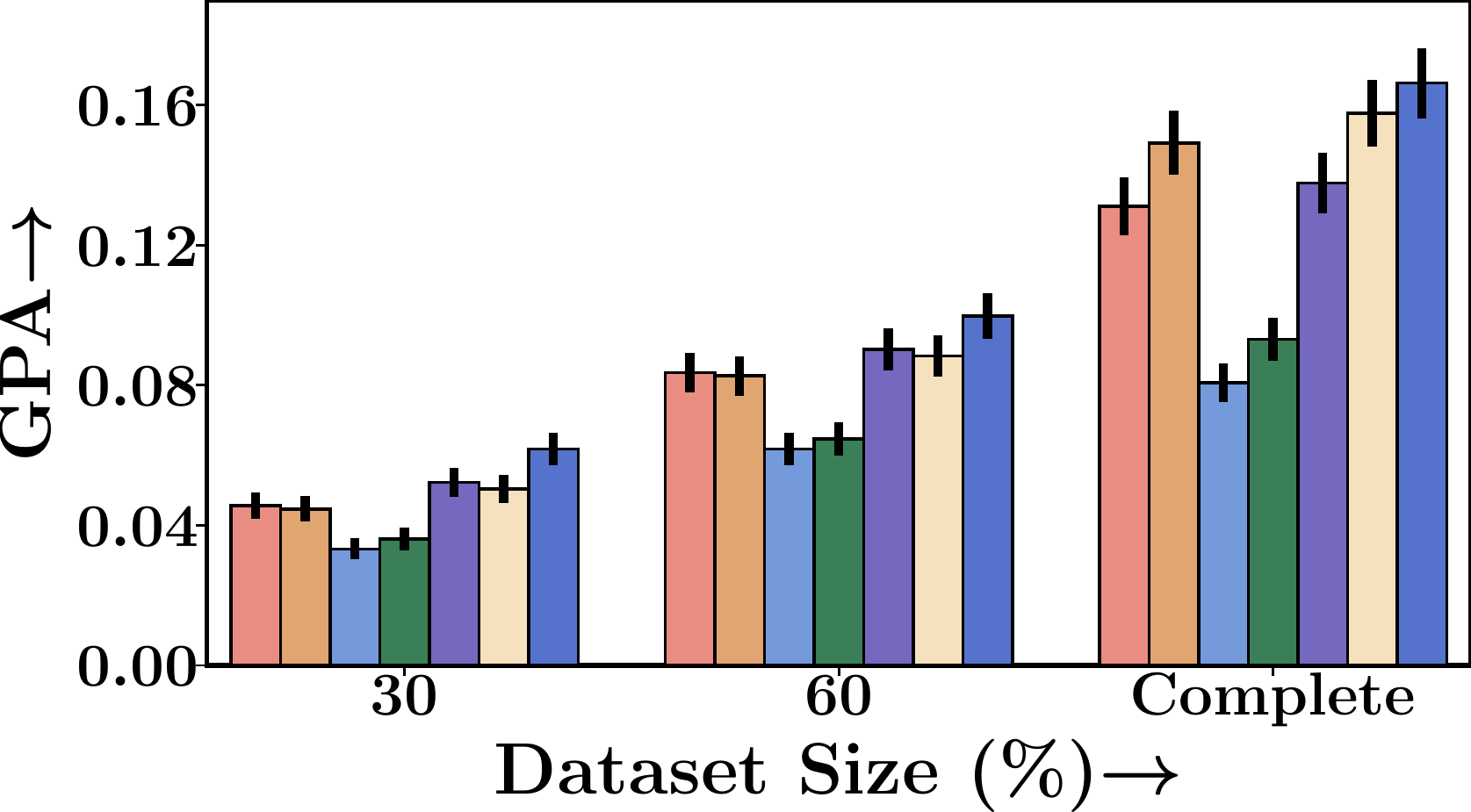}}
\caption{\mult}
\end{subfigure}
\hfill
\begin{subfigure}[b]{0.32\linewidth}
{\includegraphics[height=3.2cm]{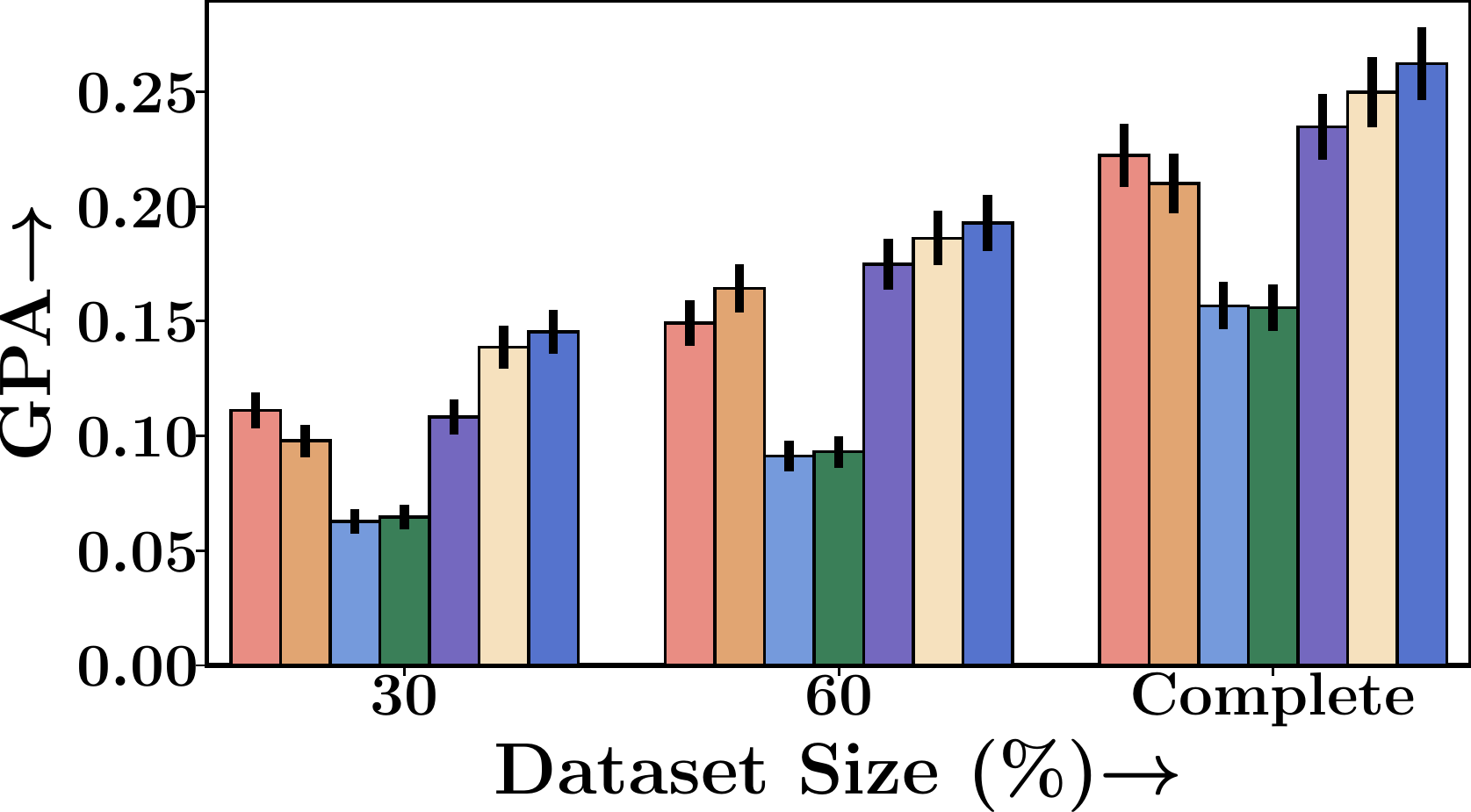}}
\caption{\act}
\end{subfigure}
{\includegraphics[height=0.4cm]{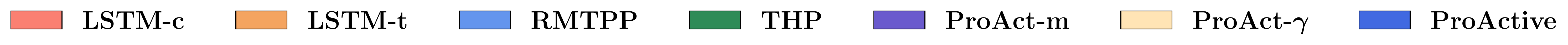}}
\vspace{-0.3cm}
\caption{Sequence goal prediction performance of \ourm, its variants -- \ours-m and \ours-$\gamma$, and other baseline models. The results show that \ourm can effectively detect the CTAS goal even with smaller test sequences as input.}
\label{fig:UU}
\end{figure*}

\subsection{Action Prediction (RQ1)}
We report on the performance of action prediction of different methods across all our datasets in Table~\ref{tab:main}. In addition, we include two variants of \ourm -- (i) \ours-c, represents our model without the goal-action hierarchy loss (Eqn. 17) and cluster-based flows (Eqn. 11 and 12), and (ii) \ours-t, represents our model without cluster-based flows. From Table~\ref{tab:main}, we note the following:
\begin{asparaitem}[$\bullet$]
\item \ourm consistently yields the best prediction performance on all the datasets. In particular, it improves over the strongest baselines by 8-27\% for time prediction and by 2-7\% for action prediction. These results signify the drawbacks of using standard sequence approaches for modeling a temporal action sequence. 

\item RMTPP~\cite{rmtpp} is the second-best performer in terms of MAE of time prediction in almost all the datasets. We also note that for \act dataset, THP~\cite{thp} outperforms RMTPP for action category prediction. However, \ourm still significantly outperforms these models across all metrics.

\item Neural MTPP methods that deploy a self-attention for modeling the distribution of action -- namely THP, SAHP, and \ourm, achieve better performance in terms of category prediction.

\item Despite AVAE~\cite{avae} being a sequence model designed specifically for activity sequences, other neural methods that incorporate complex structures using self-attention or normalizing flows easily outperform it.
\end{asparaitem}
\noindent To sum up, our empirical analysis suggests that \ourm can better model the underlying dynamics of a CTAS as compared to all other baseline models. Moreover, the performance gain over \ours-c and \ours-t highlights the need for modeling action hierarchy and cluster-based flows. 

\xhdr{Qualitative Assessment}
We also perform a qualitative analysis to highlight the ability of \ourm for modeling the inter-arrival times for action prediction. Specifically, we plot the actual inter-action time differences and the time-difference predicted by \ourm in Figure~\ref{fig:qualitative} for \mult\ and \act\ datasets. From the results, we note that the predicted inter-arrival times closely match with the true inter-arrival times and \ourm is even able to capture large time differences (peaks). For brevity, we omitted the results for \act\ dataset.

\subsection{Goal Prediction (RQ2)}
Here, we evaluate the goal detection performance of \ourm along with other baselines. To highlight the \textit{early} goal detection ability of our model,  we report the results across different variants of the test set, \ie, with the initial 30\% and 60\% of the actions in the CTAS in terms of goal prediction accuracy (GPA). In addition, we introduce two novel baselines, LSTM-c, and LSTM-t, that detect the CTAS goal using just the types and the times of actions respectively. We also compare with the two best-performing MTPP baselines -- RMTPP and THP which we extend for the task of goal detection by a $k$-means clustering algorithm. In detail, we obtain the sequence embedding, say $\bs{s}_k$ using the MTPP models and then cluster them into $|\cm{G}|$ clusters based on their cosine similarities and perform a maximum \textit{polling} across each cluster, \ie, predict the most common goal for each cluster as the goal for all CTAS in the same cluster. In addition, we introduce two new variants of our model to analyze the benefits of early goal detection procedures in \ourm -- (i) \ourm-m, represents our model without the goal-based margin loss given in Eqn.~\ref{eqn: margin} and (ii) \ourm-$\gamma$, is our model without the discount-factor weight in Eqn.~\ref{eqn: discount}. We also report the results for the complete model \ourm.

The results for goal detection in Figure \ref{fig:UU}, show that the complete design of \ourm achieves the best performance among all other models. We also note that the performance of MTPP based models deteriorates significantly for this new task which shows the unilateral nature of the prediction prowess of MTPP models, unlike \ourm. Interestingly, the variant of \ourm without the margin loss \ourm-m performs poorly as compared to the one without the discount-factor, \ourm-$\gamma$. This could be attributed to better convergence guarantees with a margin-based loss over the latter. Finally, we observe that standard LSTM models are easily outperformed by our model, thus reinforcing the need for joint training of types and action times.

\begin{figure}[t]
\centering
\begin{subfigure}[b]{0.30\columnwidth}
\includegraphics[height=2.4cm]{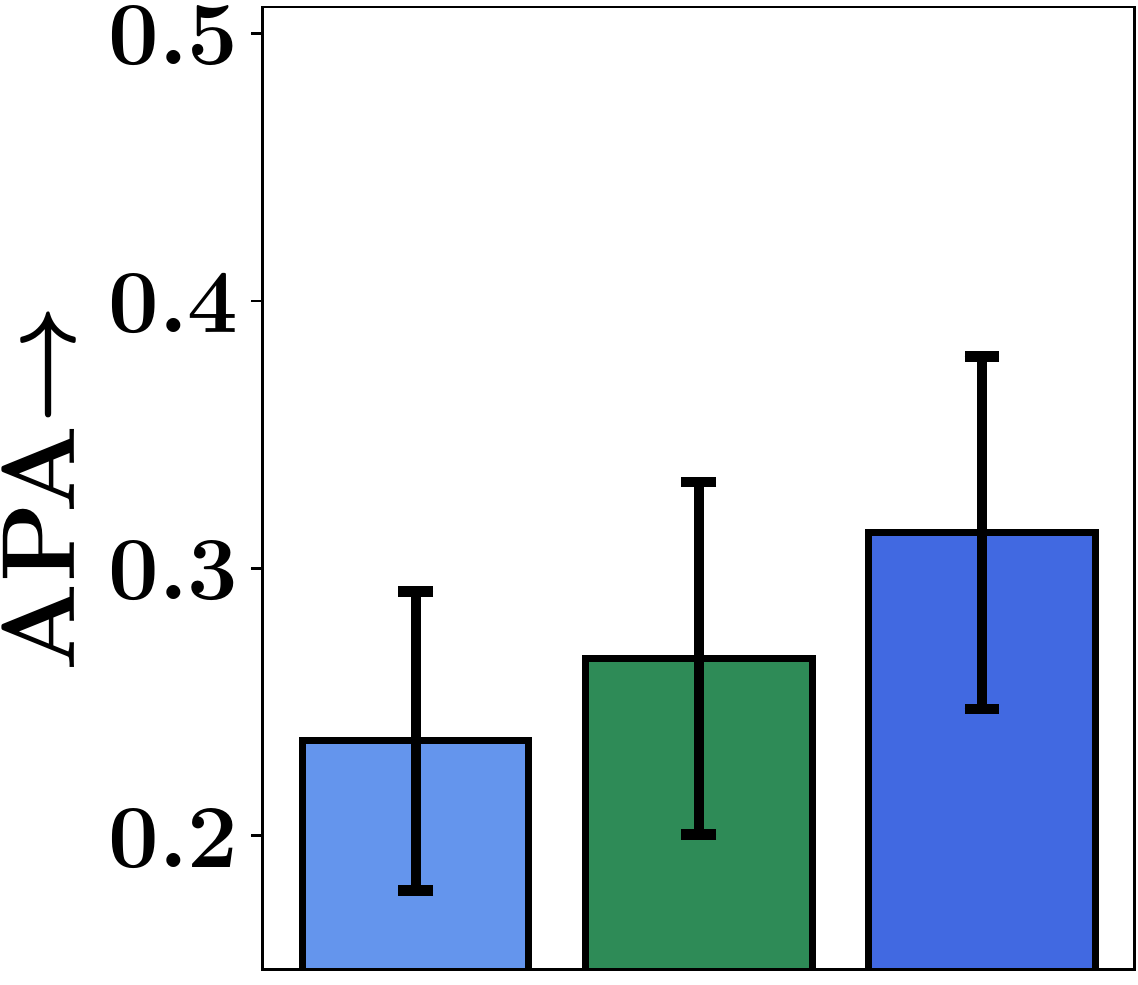}
\caption{\bfast}
\end{subfigure}
\hfill
\begin{subfigure}[b]{0.30\columnwidth}
\includegraphics[height=2.4cm]{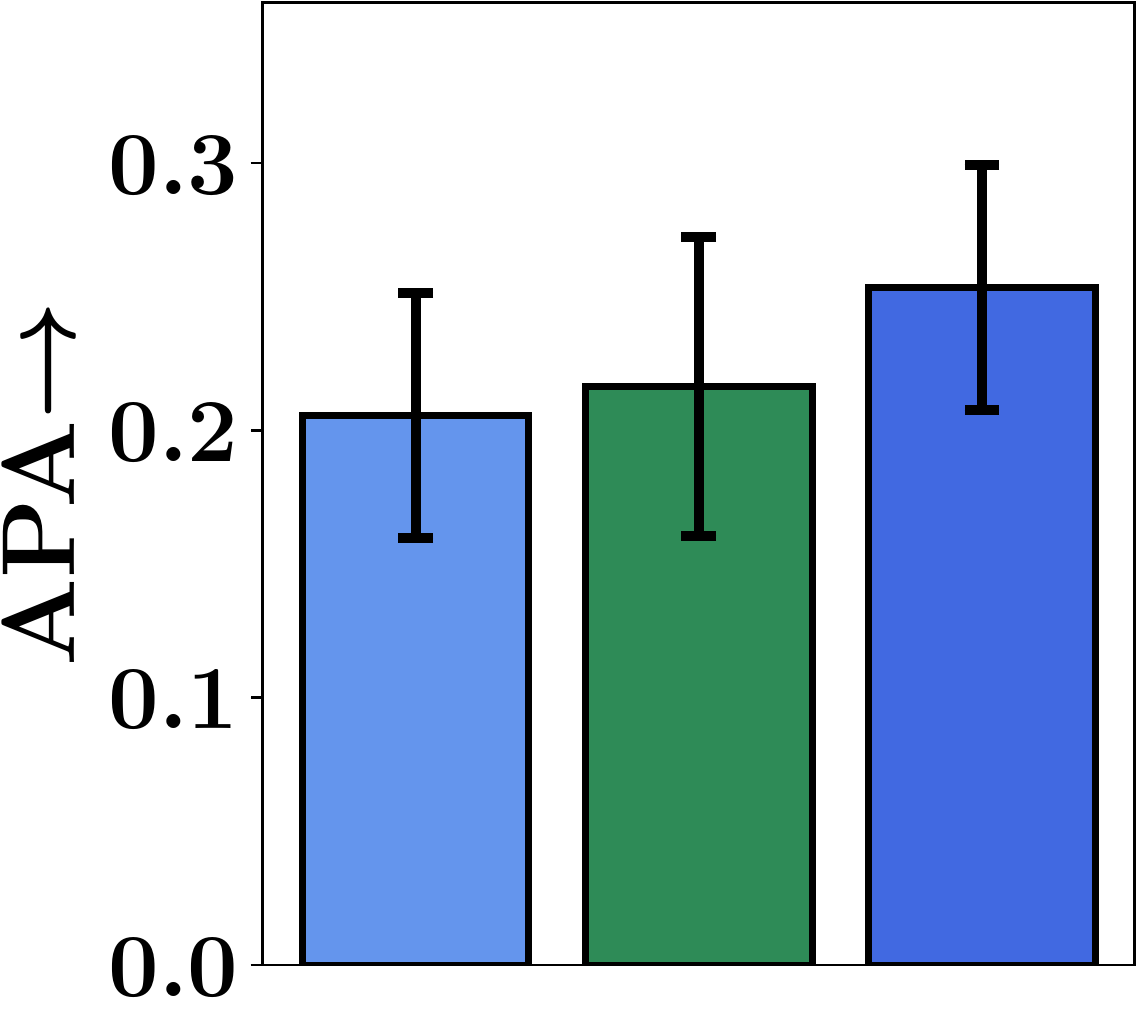}
\caption{\mult}
\end{subfigure}
\hfill
\begin{subfigure}[b]{0.30\columnwidth}
\includegraphics[height=2.4cm]{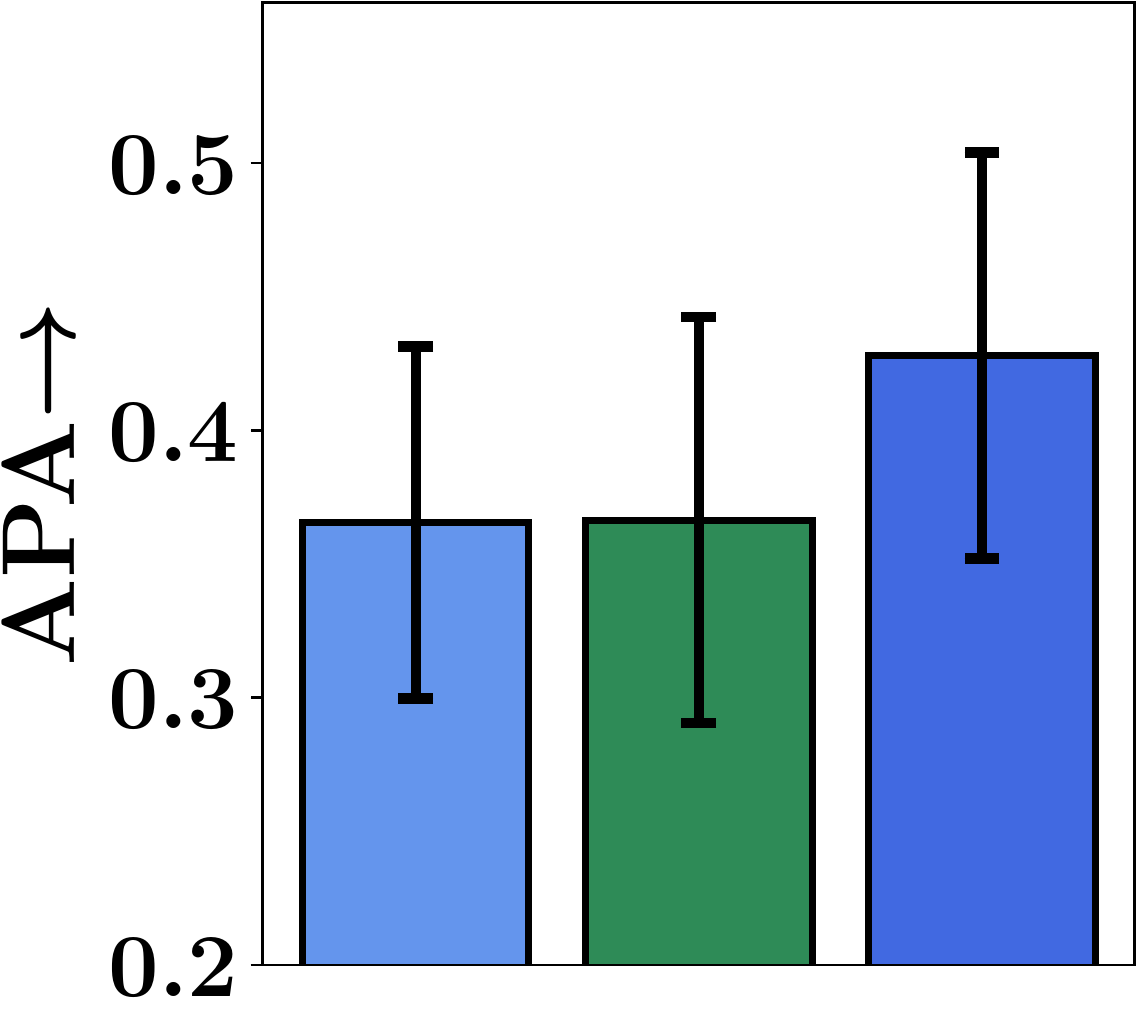}
\caption{\act}
\end{subfigure}
{\includegraphics[height=0.4cm]{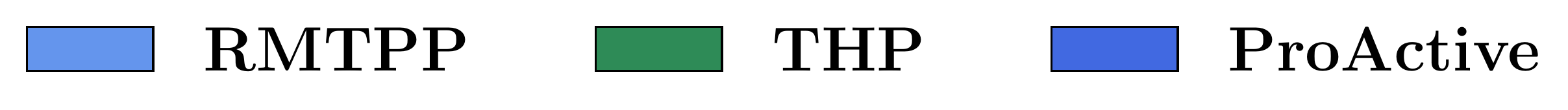}}
\vspace{-0.3cm}
\caption{\label{fig:mpa} Sequence Generation results for \ourm and other baselines in terms of APA for action prediction. \vspace{0.3cm}}
\end{figure}

\begin{figure}[t]
\centering
\begin{subfigure}[b]{0.30\columnwidth}
\includegraphics[height=2.4cm]{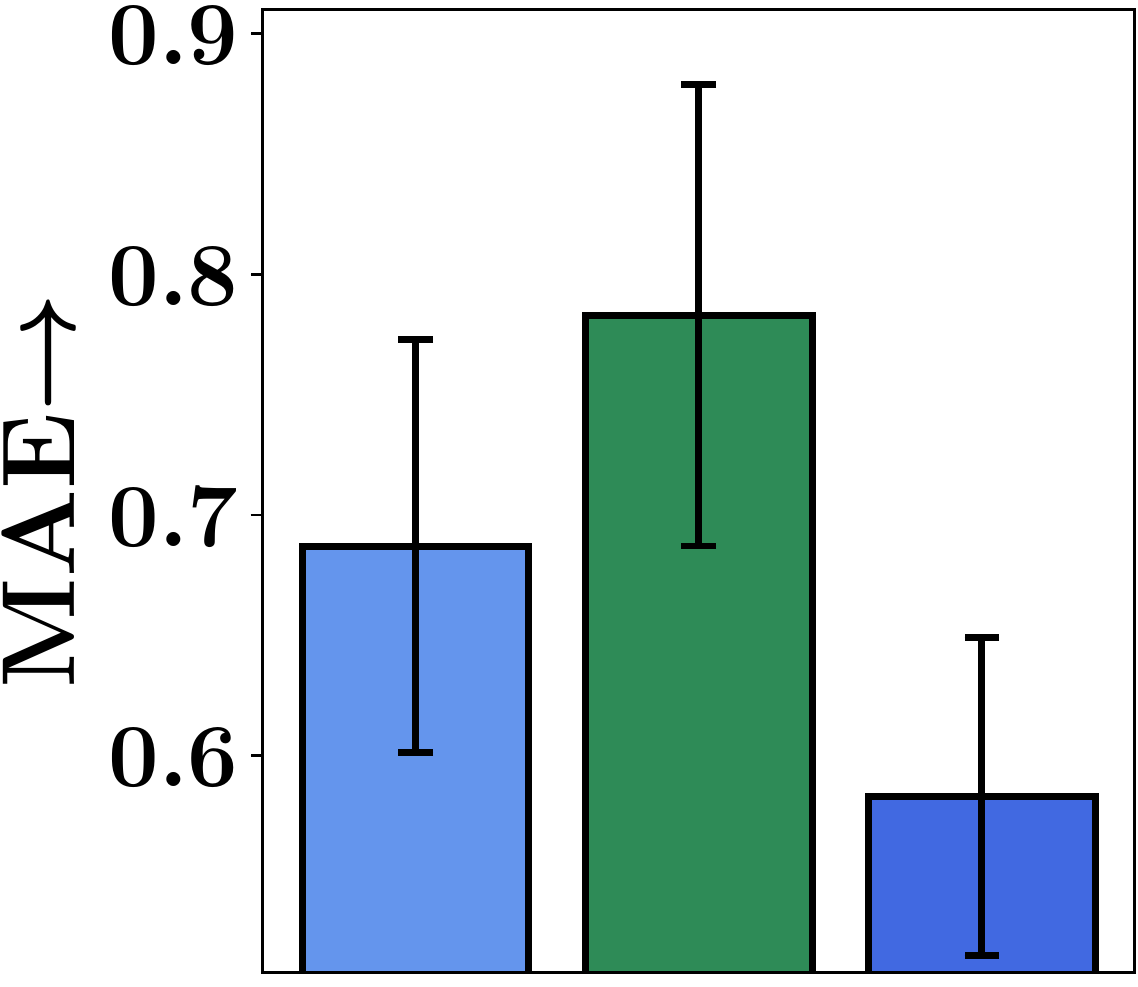}
\caption{\bfast}
\end{subfigure}
\hfill
\begin{subfigure}[b]{0.33\columnwidth}
\includegraphics[height=2.4cm]{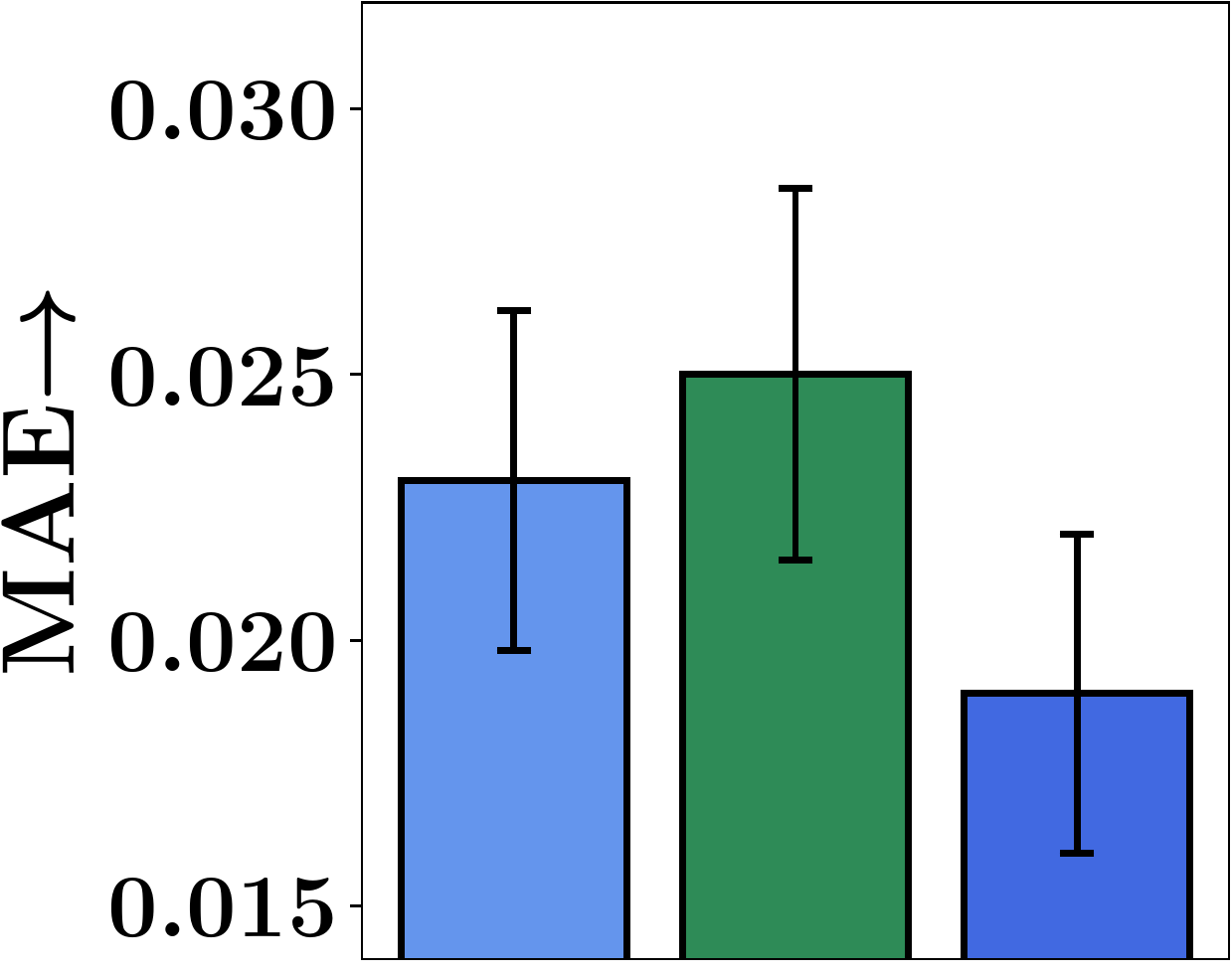}
\caption{\mult}
\end{subfigure}
\hfill
\begin{subfigure}[b]{0.30\columnwidth}
\includegraphics[height=2.4cm]{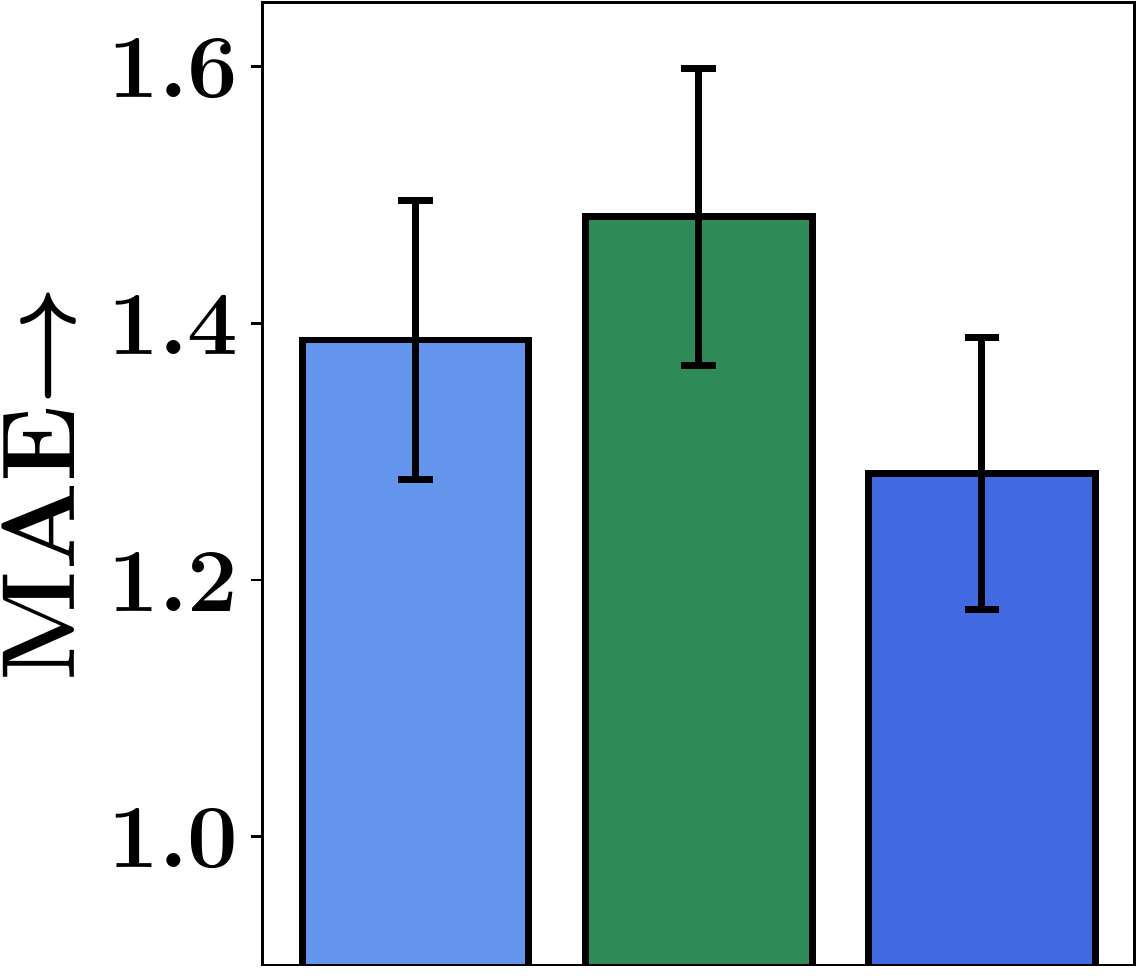}
\caption{\act}
\end{subfigure}
{\includegraphics[height=0.4cm]{figs/legend_cat.pdf}}
\vspace{-0.3cm}
\caption{\label{fig:mae} Sequence Generation results for \ourm and other baselines in terms of MAE for time prediction.}
\end{figure}

\subsection{Sequence Generation (RQ3)}
Here, we evaluate the sequence generation ability of \ourm. Specifically, we generate all the sequences in the test set by giving the \textit{true} goal of the CTAS and the first action as input to the procedure described in Section~\ref{sec:generation}. However, there may be differences in lengths of the generated and true sequences, \ie, the length of generated sequences is usually greater than the true CTAS. Therefore, we compare the actions in the true sequence with the initial $|\cm{S}|$ generated actions. Such an evaluation procedure provides us the flexibility of comparing with other MTPP models such as RMTPP~\cite{rmtpp} and THP~\cite{thp}. As these models cannot be used for end-to-end sequence generation, we alter their underlying model for \textit{forecasting} future actions given the first action and then iteratively update and sample from the MTPP parameters. We report the results in terms of APA and MAE for action and time prediction in Figure~\ref{fig:mpa} and Figure~\ref{fig:mae} respectively. The results show that \ourm can better capture the generative dynamics of a CTAS in comparison to other MTPP models. We also note that the prediction performance deteriorates significantly in comparison to the results given in Table~\ref{tab:main}. This could be attributed to the error that gets compounded in farther predictions made by the model. Interestingly, the performance advantage that \ourm has over THP and RMTPP is further widened during sequence generation. 

\xhdr{Length Comparison}
Here, we report the results for length comparison of the generated sequence and the true sequence. Specifically, we identify the count of instances where \ourm was able to effectively capture the generative mechanism of a sequence as:
\begin{equation}
\mathrm{CL} = \frac{1}{N}\sum_{\forall \cm{S}} \#(|\cm{S}|=|\widehat{\cm{S}}|),
\end{equation}
where CL denotes the \textit{Correct-Length} ratio with values $0.21$, $0.11$, and $0.16$ for datasets \bfast, \mult, and \act\ respectively. On a coarse level these results might seem substandard, however, given the difficulty associated with the problem of sequence generation using just the CTAS goal, we believe these values are satisfactory. Moreover, we believe that the sequence generation procedure of \ourm opens up new frontiers for generating action sequences.

\begin{figure}[t]
\centering
\begin{subfigure}[b]{0.32\columnwidth}
\includegraphics[height=2.5cm]{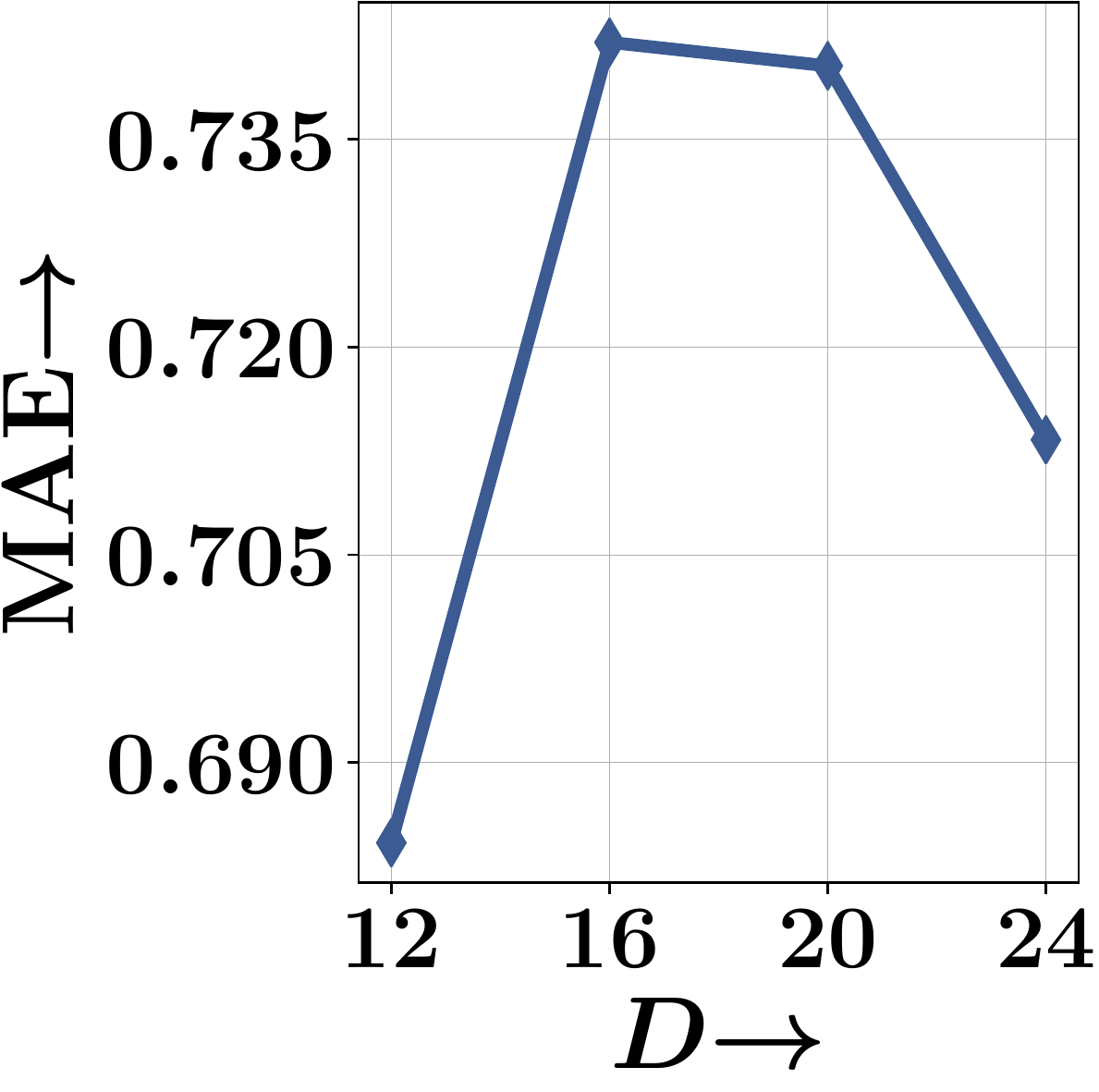}
\caption{Across $D$}
\end{subfigure}
\hfill
\begin{subfigure}[b]{0.32\columnwidth}
\includegraphics[height=2.5cm]{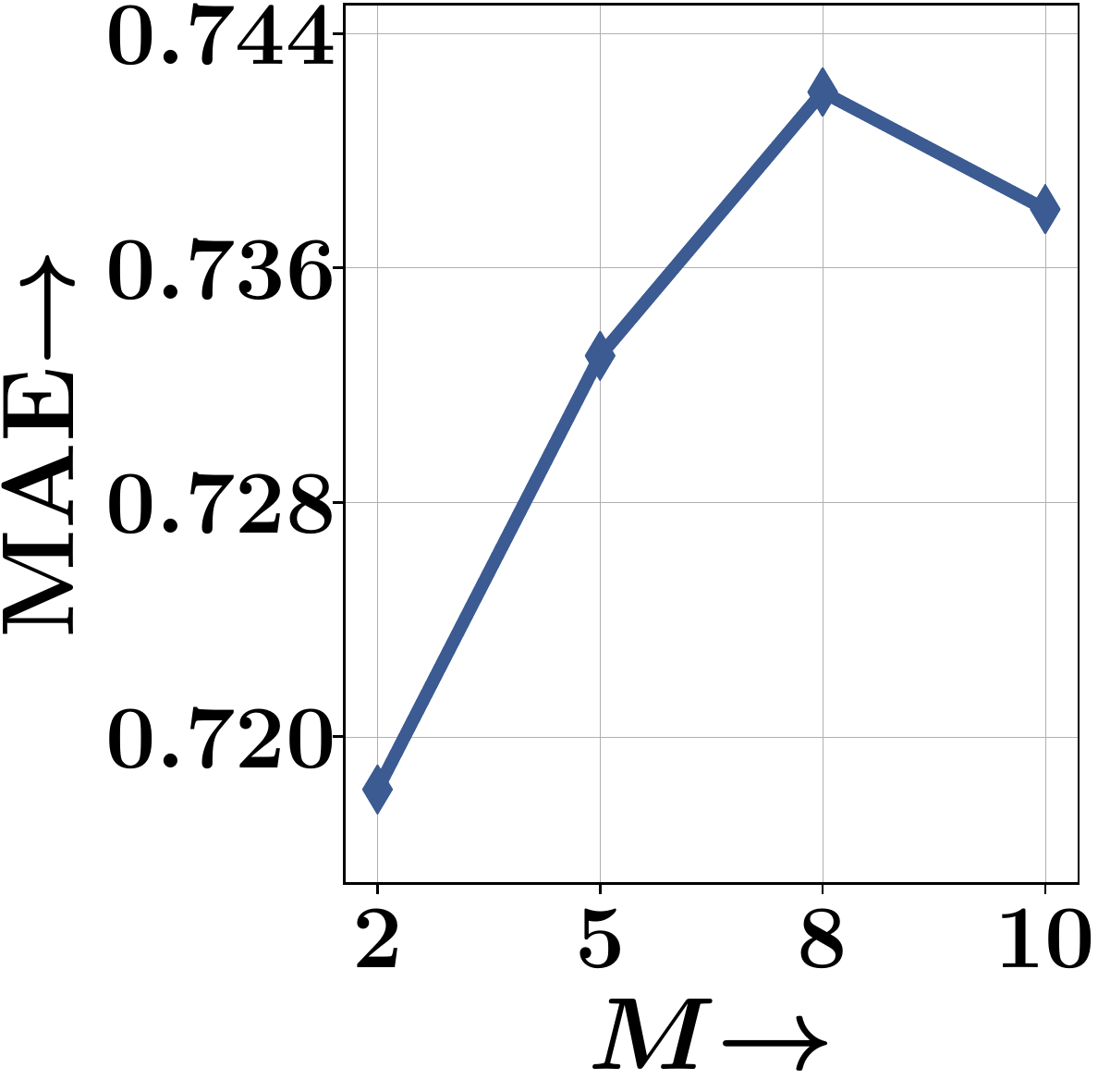}
\caption{Across $\cm{M}$}
\end{subfigure}
\hfill
\begin{subfigure}[b]{0.32\columnwidth}
\includegraphics[height=2.5cm]{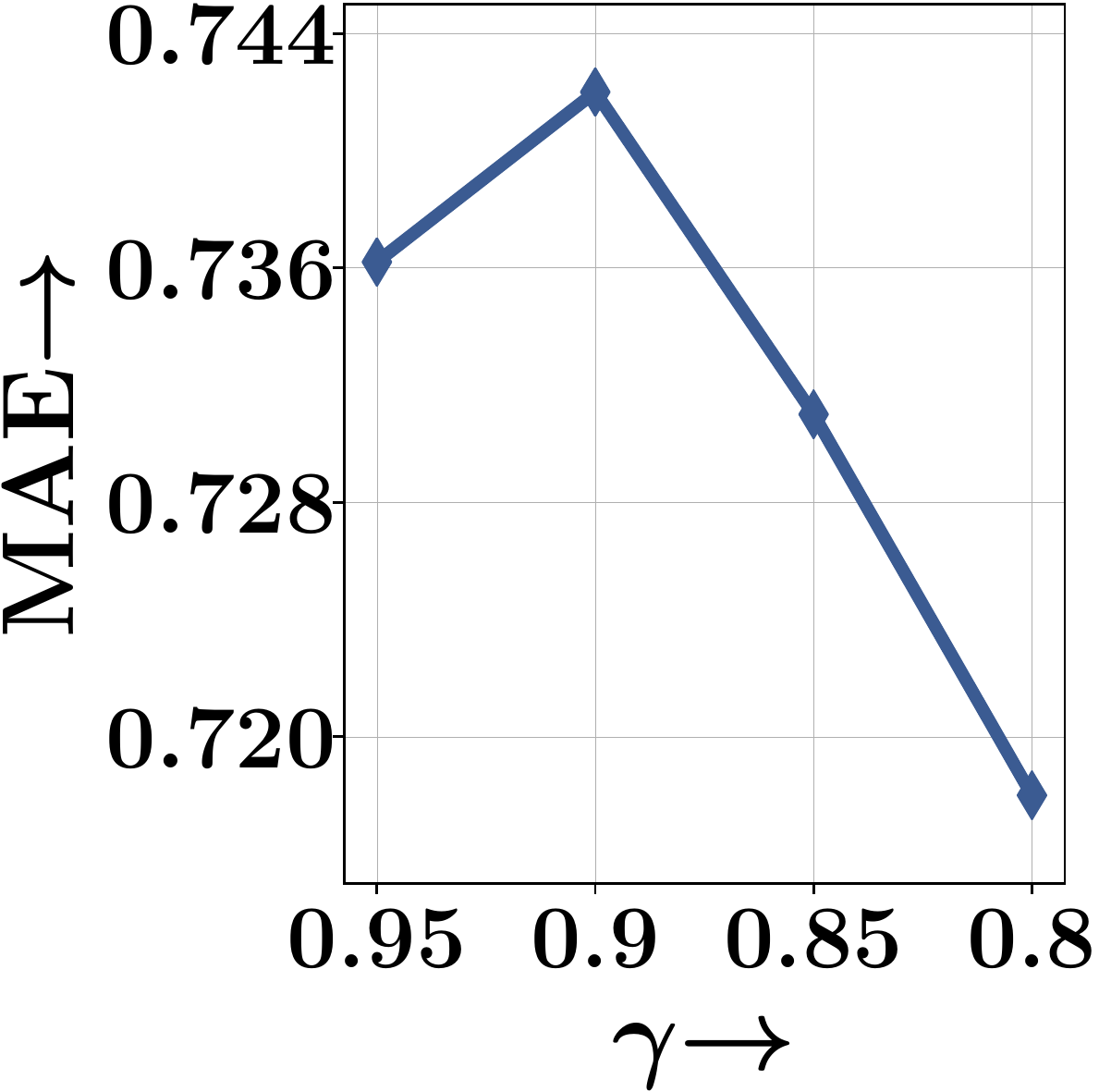}
\caption{Across $\gamma$}
\end{subfigure}
\vspace{-0.3cm}
\caption{\label{fig:pars} \ourm sensitivity for \act\ dataset with different hyper-parameter values.}
\end{figure}

\subsection{Parameter Sensitivity (RQ4)}
Finally, we perform the sensitivity analysis of \ourm over key parameters: (i) $D$, the dimension of embeddings; (ii) $\cm{M}$, no. clusters for log-normal flow; and (iii) $\gamma$, the discount factor described in Eqn.\ref{eqn: discount}. For brevity purposes, we only report results on Activity dataset, but other datasets displayed a similar behavior. From Figure \ref{fig:pars} we show the performance of \ourm across different hyper-parameter values. We note that as we increase the embedding dimension the performance first increases since it leads to better modeling. However, beyond a point the complexity of the model increases requiring more training to achieve good results, and hence we see its performance deteriorating. We see a similar trend for $\cm{M}$, as increasing the number of clusters leads to better results before saturating at a certain point. We found $\cm{M} = 5$ to be the optimal point across datasets in our experiments. Finally across $\gamma$, we notice that smaller values for gamma penalize the loss function for detecting the goal late, however, it deteriorates the action prediction performance of \ourm. Therefore, we found $\gamma = 0.9$ as the best trade-off between goal and action prediction.

\xhdr{Scalability} For all datasets, the runtimes for training \ourm are within 1 hour and thus are within the practical range for deployment in real-world scenarios. These running times further reinforce our design choice of using a neural MTPP due to their faster learning and closed-form sampling~\cite{intfree,ppflows}.

\section{Related Work} \label{sec:related}
In this section, we introduce key related work for this paper. It mainly falls into -- activity prediction and temporal point processes.

\subsection{Activity Prediction}
Activity modeling in videos is a widely used application with recent approaches focusing on frame-based prediction. \citet{conc1} predicts the future actions via hierarchical representations of short clips, \citet{conc2} jointly predicts future activity ad the starting time by capturing different sequence features and a similar procedure is adopted by \cite{conc3} that predicts the action categories of a sequence of future activities as well as their starting and ending time. \citet{margin} propose a method for early classification of a sequence of frames extracted from a video by maximizing the margin-based loss between the correct and the incorrect categories, however, it is limited to visual data and cannot incorporate the action-times. This limits its ability for use in CTAS, and sequence generation. A recent approach~\cite{avae} proposed to model the dynamics of action sequences using a variational auto-encoder built on top of a temporal point process. We consider their work as most relevant to \ourm as it also addressed the problem of CTAS modeling. However, as shown in our experiments \ourm was able to easily outperform it across all metrics. This could be attributed to the limited modeling capacity of VAE over normalizing flows. Moreover, their sampling procedure could not be extended to sequence generation. Therefore, in contrast to the past literature, \ourm is the first application of MTPP models for CTAS modeling and end-to-end sequence generation.

\subsection{Temporal Point Processes}
In recent years neural Marked Temporal Point Processes (MTPP) have shown a significant promise in modeling a variety of continuous-time sequences in healthcare~\cite{rizoiu2,neuroseqret}, finance~\cite{sahp,bacry}, education~\cite{sahebi}, and social networks~\cite{leskovec,nhp,thp,imtpp,reformd,ank}. However, due to the limitations of traditional MTPP models, in recent years, neural enhancements to MTPP models have significantly enhanced the predictive power of these models. Specifically, they combine the continuous-time approach from the point process with deep learning approaches and thus, can better capture complex relationships between events. The most popular approaches~\cite{rmtpp,nhp,intfree,sahp,thp} use different methods to model the time- and mark-distribution via neural networks. Specifically,~\citet{rmtpp} embeds the event history to a vector representation via a recurrent encoder that updates its state after parsing each event in a sequence;~\citet{nhp} modified the LSTM architecture to employ a continuous-time state evolution;~\citet{intfree} replaced the intensity function with a mixture of \textit{log-normal} flows for closed-form sampling;~\citet{sahp} utilized the transformer architecture~\citet{transformer} to capture the long-term dependencies between events in the history embedding and~\citet{thp} used the transformer architecture for sequence embedding but extended it to graph settings as well. However, these models are not designed to capture the generative distribution of future events in human-generated sequences.

\section{Conclusion} \label{sec:conc}
Standard deep-learning models are not designed for modeling sequences of actions localized in continuous time. However, neural MTPP models overcome this drawback but have limited ability to model the events performed by a human. Therefore, in this paper, we developed a novel point process flow-based architecture called \ourm for modeling the dynamics of a CTAS. \ourm solves the problems associated with action prediction, goal prediction, and for the first time, we extend MTPP for end-to-end CTAS generation. Our experiments on three large-scale diverse datasets reveal that \ourm can significantly improve over the state-of-the-art baselines across all metrics. Moreover, the results also reinforce the novel ability of \ourm to generate a CTAS. We hope that such an application will open many horizons for using MTPP in a wide range of tasks. As a future work, we plan to incorporate a generative adversarial network~\cite{gan,wgantpp} with action sampling and train the generator and the MTPP model simultaneously.

\begin{acks}
This work was partially supported by a DS Chair of AI fellowship to Srikanta Bedathur. 
\end{acks}

\bibliographystyle{ACM-Reference-Format}
\bibliography{refs}
\end{document}